\pdfoutput=1

\documentclass[11pt]{article}

\usepackage[preprint]{acl}

\usepackage{times}
\usepackage{latexsym}

\usepackage[T1]{fontenc}

\usepackage[utf8]{inputenc}

\usepackage{microtype}

\usepackage{inconsolata}

\usepackage{graphicx}

\usepackage{booktabs}
\usepackage{hyperref}
\usepackage{amsfonts}
\usepackage{amssymb}
\usepackage{amsmath}
\usepackage[utf8]{inputenc}
\usepackage{textcomp}
\usepackage{newunicodechar}
\newunicodechar{−}{\textminus}

\usepackage{tabularx}
\usepackage{array}    
\usepackage{rotating} 

\usepackage{amssymb}
\usepackage{multirow}
\usepackage{booktabs}
\usepackage{amsmath} 
\usepackage[table]{xcolor}

\usepackage{colortbl}
\usepackage{xcolor}
\usepackage[normalem]{ulem}
\usepackage{pifont}

\title{Enhancing Target-Guided Proactive Dialogue Systems via Conversational Scenario Modeling and Intent-Keyword Bridging
}

\author{
 \textbf{Maodong Li\textsuperscript{1,2}},
 \textbf{Yancui Li\textsuperscript{3}},
 \textbf{Fang Kong\textsuperscript{1,2}}\thanks{Corresponding author}
\\
 \textsuperscript{1}School of Computer Science and Technology, Soochow University, China \\
 \textsuperscript{2}Jiangsu Key Lab of Language Computing, Suzhou 215123, China
 \\
 \textsuperscript{3}School of Computer and Information Engineering, Henan Normal University, China
\\
 \texttt{\{20254027002@stu,kongfang@\}suda.edu.cn} \quad
 \texttt{liyancui@htu.edu.cn}
}

\begin{document}
\maketitle

\begin{abstract}
A target-guided proactive dialogue system aims to steer conversations proactively toward pre-defined targets, such as designated keywords or specific topics. During guided conversations, dynamically modeling conversational scenarios and intent keywords to guide system utterance generation is beneficial; however, existing work largely overlooks this aspect, resulting in a mismatch with the dynamics of real-world conversations. In this paper, we jointly model user profiles and domain knowledge as conversational scenarios to introduce a scenario bias that dynamically influences system utterances, and employ intent-keyword bridging to predict intent keywords for upcoming dialogue turns, providing higher-level and more flexible guidance. Extensive automatic and human evaluations demonstrate the effectiveness of conversational scenario modeling and intent–keyword bridging, yielding substantial improvements in proactivity, fluency, and informativeness for target-guided proactive dialogue systems, thereby narrowing the gap with real-world interactions.
\end{abstract}
\section{Introduction}
Target-guided proactive dialogue systems are designed to proactively steer conversations toward pre-defined targets, such as designated keywords or specific topics \cite{wang2024target,zhang-etal-2025-enhancing-goal,kang2026pseudo}. Compared to passively responding to users, proactive guidance better aligns with real-world interaction patterns and enhances user engagement \cite{wu-etal-2025-interpersonal}. Such systems have long been a central focus in natural language processing and have been applied across diverse domains, such as recommendation systems, emotional dialogue, and medical consultations \cite{dao-etal-2024-experience,xu-etal-2024-reasoning,hao-kong-2025-enhancing,wu-etal-2025-personas}.

\begin{figure}[t]
  \centering
  \includegraphics[width=\columnwidth]{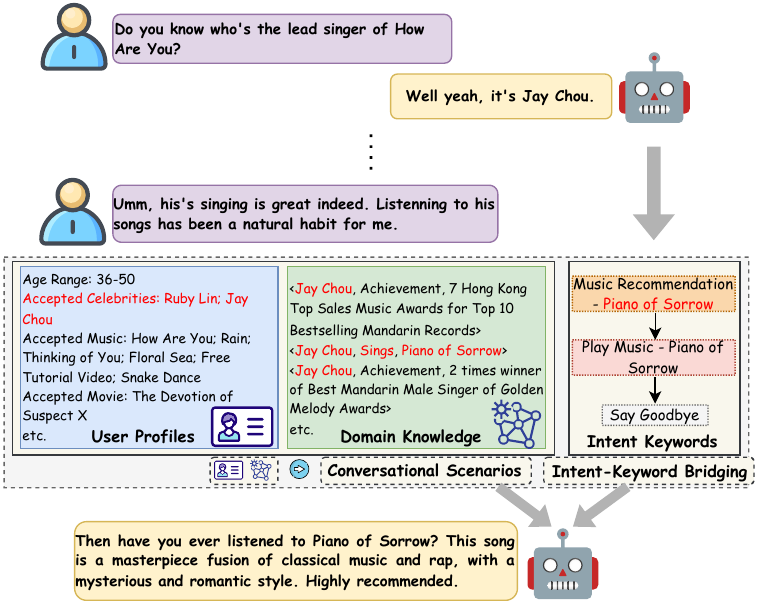}
  \caption{An example of a target-guided proactive dialogue system, where the red highlights denote snippets relevant to the current dialogue, which we analyze in detail in App.~\ref{case_study}.
  }
  \label{example_1}
\end{figure}

Figure~\ref{example_1} presents an example of a target-guided proactive dialogue system. When a user produces a new utterance, the system predicts intent keywords for the next few steps based on the dialogue context, represented as a sequence (e.g., Music Recommendation – Piano of Sorrow $\rightarrow$ Play Music – Piano of Sorrow $\rightarrow$ Say Goodbye). In this paper, we refer to these as intent keywords, which indicate the system's expected future behavior. At the same time, the system dynamically selects points of interest based on user profiles and domain knowledge, and correspondingly influences the system utterance to ensure alignment with the current context\footnote{We use the term "utterance" instead of "response" to better capture the proactive guidance in our setting.}, thereby guiding the conversation toward achieving the pre-defined target while maintaining engagement.

Most previous studies on target-guided proactive dialogue systems focus on static entity-based keyword planning \cite{tang-etal-2019-target,yang-etal-2022-topkg}. These keywords indicate the entities that the next system utterance should focus on, but they provide limited information regarding the system's expected utterance. Subsequently, \citet{wang-etal-2023-dialogue,dao-etal-2023-reinforced} employ more semantic keywords to guide the generation of system utterances, which we refer to as intent keywords, as they indicate the system's expected behavior with higher-level guidance. Although \citet{wang2023target,wang2024target} utilize intent keywords, they focus only on the next-turn intent keyword when guiding utterance generation, overlooking the fact that subsequent turns and their intent keywords are typically consistent and coherent. Considering the intent keywords of multiple subsequent turns simultaneously would provide greater flexibility. Furthermore, the system's utterances should remain consistent with user profiles and domain knowledge \cite{zhang-etal-2025-enhancing-goal}, which rely on an external model; instead, we jointly model user profiles and domain knowledge, maintaining such consistency to improve proactivity and enhance engagement.




To this end, we introduce the concept of conversational scenarios, which capture the conversational background of the current interaction, and intent–keyword bridging, which dynamically predicts intent keywords for the next few dialogue turns. As shown in Figure~\ref{example_1}, we jointly model user profiles and domain knowledge as conversational scenarios and dynamically influence system utterances, ensuring that the generated utterances remain consistent with the ongoing scenarios. Unlike static planning, our intent–keyword bridging dynamically predicts intent keywords for upcoming dialogue turns, providing higher-level and more flexible guidance. Extensive automatic and human evaluations demonstrate the effectiveness of conversational scenario modeling and intent–keyword bridging, yielding substantial improvements in proactivity, fluency, and informativeness for target-guided proactive dialogue systems. Our contributions are summarized as follows:

\begin{itemize}
	\item We present conversational scenarios by jointly modeling user profiles and domain knowledge, introducing a bias that dynamically influences system utterances, thereby enabling more precise initiative and enhancing user engagement.
	\item We propose intent–keyword bridging to dynamically predict intent keywords for upcoming dialogue turns, offering higher-level and more flexible guidance\footnote{https://github.com/imaodong/EnTarget-Guided\_Proactive\_Dialog.}.	
\end{itemize}




\section{Methodology}
Our framework is illustrated in Figure~\ref{Frame_1}. It comprises two components: \textbf{Conversational Scenario Modeling (CSM)} and \textbf{Intent-Keyword Bridging (IKB)}. In the CSM, we jointly model user profiles and domain knowledge as conversational scenarios, introducing a bias that dynamically influences system utterances. This ensures that generated utterances remain consistent with the ongoing scenarios, enabling more precise initiative and enhancing user engagement. In the IKB, intent keywords for the next few turns are dynamically predicted based on the conversational scenario and dialogue history, capturing the expected behavior in the next turn and anticipating actions several steps ahead, thereby providing higher-level and flexible guidance.

\subsection{Task Formulation and Notation}
Suppose $D = \{(r^{(i)}, h^{(i)}, g^{(i)}, S^{(i)}, Z^{(i)0:m})\}^{i=1}_{N}$ is a target-guided dialogue dataset containing $N$ samples, where $r^{(i)}$ denotes the system utterance, $h^{(i)}$ denotes the dialogue history, and $g^{(i)}$ denotes the pre-defined dialogue target. $S^{(i)} = (u^{(i)}, k^{(i)})$ represents the conversational scenario, comprising user profiles $u^{(i)}$ and domain knowledge $k^{(i)}$, and $Z^{(i)0:m}$ denotes the keyword bridging sequence, with $m$ representing the number of future-turn intent keywords determined empirically. The task is to generate system utterances $r^{(i)}$ to guide the conversation toward achieving $g^{(i)}$, while maintaining proactivity, engagement, and naturalness.

\subsection{Conversational Scenario Modeling}
\label{CSM_section}
The conversational scenario captures the conversational background of the current interaction. To our knowledge, this work is the first to jointly model user profiles and domain knowledge as conversational scenarios, as they together determine the current interaction state for both the user and the system. The conversational scenario introduces a bias that dynamically influences system utterances, enabling the system to take more precise initiative, enhancing user engagement, and ensuring that generated utterances remain consistent with the ongoing scenario. Let $\text{Enc}(\cdot)$ denote the T5 encoder \cite{chung2022scalinginstructionfinetunedlanguagemodels} backbone; the modeling process can thus be formalized as follows:
\begin{equation}
	\mathbf{b} = \text{Softmax}(\mathbf{B}\cdot(\mathcal{F}_k(\mathbf{H}^k)+ \mathcal{F}_u(\mathbf{H}^u)))
\end{equation}
\begin{equation}
    \mathbf{H}^k,\mathbf{H}^u,\mathbf{H}^h=\text{Enc}(k),\text{Enc}(u),\text{Enc}([h;g])
\end{equation}
where $\mathbf{H}^{k} \in \mathbb{R}^{l_k \times d}$, $\mathbf{H}^{u} \in \mathbb{R}^{l_u \times d}$, and $\mathbf{H}^{h} \in \mathbb{R}^{l_h \times d}$ denote the hidden states of domain knowledge, user profiles, and dialogue history, respectively. Here, $l_k$ and $l_u$ denote the lengths of the domain knowledge and user profiles, respectively, while $l_h$ denotes the length of the dialogue history and dialogue target. The variable $d$ represents the hidden dimension. $\mathcal{F}_k(\cdot)$ and $\mathcal{F}_u(\cdot)$ are mapping functions implemented using average pooling followed by a multi-layer perceptron. Then $\mathcal{F}_k(\mathbf{H}^k)$ and $\mathcal{F}_u(\mathbf{H}^u)$ jointly constitute the proposed conversational scenario bias $\mathbf{b} \in \mathbb{R}^{1 \times \mathcal{V}}$ via element-wise addition. Here, $\mathbf{B}$ denotes trainable parameters, and $\mathcal{V}$ denotes the vocabulary size.


\subsection{Intent-Keyword Bridging}
Intent–keyword bridging serves as a connection between the dialogue history and the system's future behaviors \cite{sevegnani-etal-2021-otters}, and we employ it to dynamically predict intent keywords for upcoming turns. We use intent keywords that consist of both keyword-type and keyword-topic \cite{liu-etal-2021-durecdial}. Let $A = \{a_1, a_2, \dots, a_{x_a}\}$ and $T = \{t_1, t_2, \dots, t_{x_t}\}$ denote the sets of keyword-types and keyword-topics, respectively, where $x_a$ and $x_t$ represent their corresponding cardinalities. We use $\zeta(\cdot)$ to denote the selection of the corresponding keyword-type/topic index, and the extraction of keyword-type/topic can be formalized as:
\begin{equation}
    \mathbf{E}^a,\mathbf{E}^t=\text{Emb}_a(\zeta(A)),\text{Emb}_t(\zeta(T))
\end{equation}
\begin{equation}
    A = \text{CLS}_a(\text{IF}(\textbf{H}^h,\mathcal{F}_k(\mathbf{H}^k),\mathcal{F}_u(\mathbf{H}^u)))
\end{equation}
\begin{equation}
    T = \text{CLS}_t(\text{IF}(\textbf{H}^h,\mathcal{F}_k(\mathbf{H}^k),\mathcal{F}_u(\mathbf{H}^u)))
\end{equation}
where $\mathbf{E}^a \in \mathbb{R}^{m \times d}$ and $\mathbf{E}^t \in \mathbb{R}^{m \times d}$ denote the embeddings of keyword-type and keyword-topic, respectively, and $m$ denotes the hyperparameter specifying the number of future turns to be predicted. $\text{CLS}_a(\cdot)$ and $\text{CLS}_t(\cdot)$ represent the classification heads for keyword-type and keyword-topic, respectively, and $\text{IF}(\cdot)$ denotes the information fusion mechanism, following \citet{wang2023target}. The resulting bridging intent keywords are formalized as follows:
\begin{equation}
    \mathbf{H}^z=\text{CONCAT}(\mathbf{H}^a;\mathbf{H}^t)
\end{equation}
\begin{equation}
    \mathbf{H}^a,\mathbf{H}^t=\text{MP}(\mathbf{E}^a),\text{MP}(\mathbf{E}^t)
\end{equation}
where $\mathbf{H}^{a} \in \mathbb{R}^{1 \times d}$, $\mathbf{H}^{t} \in \mathbb{R}^{1 \times d}$, and $\mathbf{H}^{z} \in \mathbb{R}^{2 \times d}$ denote the hidden states of the keyword-type, keyword-topic, and intent keyword, respectively. $\text{MP}(\cdot)$ denotes the max-pooling operation. By dynamically predicting $m$ intent keywords and applying max pooling, we obtain the most relevant bridging intent keyword $\mathbf{H}^z$ for the next turn, considering the next $m$ turns. The final step in our framework, generating the system utterance, can be represented as\footnote{Implementation details are provided in App.~\ref{imple_details}}:
\begin{equation}
	r_t=\arg\max P(r_t|r_{<t},\mathbf{H}^z,\mathbf{H}^h,\mathbf{b})
\end{equation}



\begin{figure}[t]
  \centering
  \includegraphics[width=\linewidth]{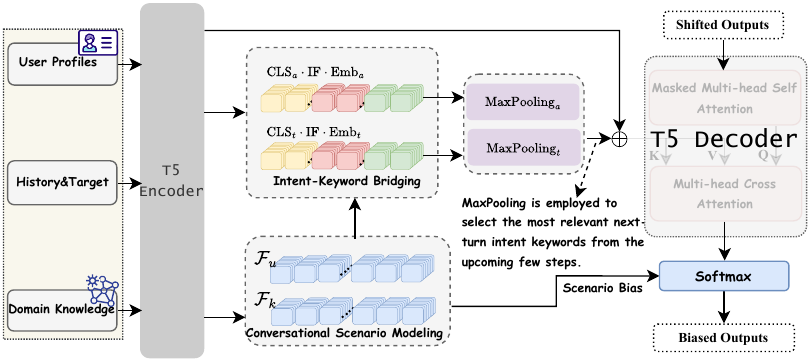}
  \caption{Our framework.}
  \label{Frame_1}
\end{figure}

\subsection{Learning Objectives}
The first learning objective is to predict the intent keywords for the next $m$ turns. We minimize the following negative log-likelihood (NLL):

\begin{equation}
    \mathcal{L}_{cls}=-\frac{1}{N}\sum_{i=1}^{N}\sum_{j=1}^{|V|} y_j^{z(i)} \log p_{\theta}(j|S^{(i)},h^{(i)},g^{(i)})
\end{equation}
where $\theta$ denotes the model parameters, and $y^{z(i)} \in \{0,1\}^{|V|}$ is a multi-hot vector with $|V| = x_a x_b$ entries, among which $2m$ entries are set to 1.0, corresponding to the ground-truth intent keywords. For generation, the probability distribution over the vocabulary at step $t$ is computed as $r_t^{(i)} = \text{softmax}(\mathbf{W}\mathbf{H}_t^{\text{last}} + \mathbf{b})$, where $\mathbf{H}_t^{\text{last}}$ denotes the final hidden state of the T5 decoder at step $t$. Let $c^{(i)} = (S^{(i)}, h^{(i)}, g^{(i)})$. We jointly optimize the conversational scenarios by minimizing the following NLL over the target utterance of length $T_i$:

\begin{equation}
	\mathcal{L}_{lm} = -\frac{1}{N} \sum_{i=1}^{N} \sum_{t=1}^{T_i} \log p_\theta(y_t^{(i)} \mid y_{<t}^{(i)}, c^{(i)}, Z^{(i)})
\end{equation}
Overall, the training objectives are:
\begin{equation}
    \mathcal{L}_{total} = \mathcal{L}_{lm} + \mathcal{L}_{cls}
\end{equation}

\subsection{Hard and Soft Modes}
\label{hard_soft_modes}
Our framework operates in two modes: hard and soft. In the hard mode, the framework predicts the intent keywords for the next $m$ turns during training and selects the top $m$ intent keywords during inference. In contrast, the soft mode predicts intent keywords for the next $m$ turns during training, but during inference, it selects intent keywords only when their probability exceeds a threshold $\delta$, thereby accounting for prediction uncertainty. For simplicity, $\text{Emb}_a(\cdot)$ and $\text{Emb}_t(\cdot)$ are collectively denoted as $\text{Emb}_z(\cdot)$. In the hard mode, the $m$ intent keywords with the highest probabilities are selected:
\begin{equation}
	\mathbf{E}^z = \text{Emb}_z(\text{Top}_m(P_z))
\end{equation}
where $\mathbf{E}^z \in \mathbb{R}^{m \times d}$ denotes the embeddings of intent keywords, and $P_z$ denotes the probability distribution over intent keywords predicted by the classification head. $\text{Top}_m(\cdot)$ selects the $m$ most probable intent keywords. However, this hard mode does not account for prediction uncertainty. In the soft mode, we select the intent keywords whose probabilities exceed a threshold $\delta$, yielding a dynamic index set $\mathcal{I}_{\delta} = \{j \mid P_z(j) \ge \delta \}$. The embeddings are then weighted by their corresponding probabilities:
\begin{equation}
	\mathbf{E}^z_j = P_z(j) \cdot \text{Emb}_z(j), \quad \forall j \in \mathcal{I}_{\delta}
\end{equation}


\begin{table*}[!t]
	\small
	\centering
	\setlength{\tabcolsep}{0.2mm}{
	\begin{tabular}{lcccccccccccc}
		\toprule
		\multicolumn{13}{c}{\textbf{DuRecDial Dataset}}                                                                                                                             \\ \cmidrule(l){2-13} 
		\multirow{2}{*}{\textbf{Model}} & \multicolumn{6}{c|}{\textbf{ID Test Set}}                                              & \multicolumn{6}{c}{\textbf{OOD Test Set}}                          \\ \cmidrule(l){2-13} 
		& \textbf{PPL$_\downarrow$}  & \textbf{W. F1}  & \textbf{BLEU-1/2}       & \textbf{DIST-1/2}       & \textbf{K. F1}  & \multicolumn{1}{c|}{\textbf{Fail.$_\downarrow$}} & \textbf{PPL$_\downarrow$}  & \textbf{W. F1}  & \textbf{BLEU-1/2}        & \textbf{DIST-1/2}       & \textbf{K. F1}  & \textbf{Fail.$_\downarrow$} \\ \midrule
		\textbf{TPDial}$^\diamond$             & \textbf{3.31} & 40.77 & 0.387/0.312 & \textbf{0.012}/\textbf{0.066} & \underline{54.08} & \multicolumn{1}{c|}{23.19} & \underline{4.18} & 34.95 & 0.348/0.271  & \textbf{0.013}/0.058 & 34.80 & 80.05 \\
		\textbf{TRIPDial}$^\diamond$              & 3.46 & 44.24 & 0.404/0.327 & \textbf{0.012}/\textbf{0.066} & \textbf{54.64} & \multicolumn{1}{c|}{22.60} & \textbf{4.17} & 42.68 & 0.411/0.335  & \textbf{0.013}/0.059 & \underline{47.13} & 22.19 \\
		\textbf{T5-Zh}                  & \underline{3.37} & 44.35 & \underline{0.410}/0.330 & \underline{0.011}/\underline{0.064} & 49.35 & \multicolumn{1}{c|}{\underline{22.30}} & 4.56 & 40.70 & 0.394/0.312  & \underline{0.012}/0.059 & 41.42 & 23.44 \\
		\rowcolor[gray]{0.9}
		\textbf{Our (hard)}              & 3.89 & \underline{44.70} & \textbf{0.433}$^\dagger$/\underline{0.347}$^\dagger$ & 0.010/0.061 & 49.62 & \multicolumn{1}{c|}{\textbf{17.13}$^\dagger$} & 5.36 & \underline{44.20}$^\dagger$ & \underline{0.431}$^\dagger$/\underline{0.347}$^\dagger$  & \underline{0.012}/\textbf{0.066}$^\dagger$ & \textbf{47.83} & \textbf{19.70}$^\ddagger$ \\
		\rowcolor[gray]{0.9}
		\textbf{Our (soft)}              & 3.87 & \textbf{44.87} & \textbf{0.433}$^\dagger$/\textbf{0.348}$^\dagger$ & 0.010/0.061 & 49.83 & \multicolumn{1}{c|}{\textbf{17.13}$^\dagger$} & 5.34 & \textbf{44.33}$^\dagger$ & \textbf{0.432}$^\dagger$/\textbf{0.348}$^\dagger$  & \underline{0.012}/\underline{0.065}$^\dagger$ & \textbf{47.83} & \underline{20.20} \\ \midrule
		\multicolumn{13}{c}{\textbf{DuRecDial2.0   Dataset}}                                                                                                                        \\ \midrule
		\textbf{TPDial}$^\diamond$             & 5.23 & 38.29 & 0.320/0.222 & \textbf{0.024}/0.083 & 50.97 & \multicolumn{1}{c|}{63.88} & 7.67 & 31.71 & 0.273/0.178 & \textbf{0.023}/0.078 & 16.02 & 99.69  \\
		\textbf{TRIPDial}$^\diamond$              & 6.19 & 35.06 & 0.313/0.223 & \underline{0.022}/0.074 & 43.35 & \multicolumn{1}{c|}{67.11} & 7.85 & 30.18 & 0.272/0.182  & 0.022/0.070 & 30.17 & 73.21 \\
		\textbf{T5-Flan}                & \underline{3.96} & 42.72 & \underline{0.390}/0.288 & 0.018/0.071 & 55.26 & \multicolumn{1}{c|}{\underline{21.29}} & \textbf{6.46} & 37.46 & 0.339/0.235  & 0.015/0.053 & 41.21 & \textbf{22.12} \\
		\rowcolor[gray]{0.9}
		\textbf{Our (hard)}              & \underline{3.96} & \underline{44.79}$^\dagger$ & \textbf{0.416}$^\dagger$/\underline{0.309}$^\dagger$ & \underline{0.022}/\textbf{0.088}$^\dagger$ & \underline{61.11}$^\dagger$ & \multicolumn{1}{c|}{\textbf{19.77}} & 7.27 & \underline{39.61}$^\dagger$ & \underline{0.363}$^\dagger$/\underline{0.252}$^\dagger$  & \underline{0.022}/\textbf{0.082}$^\dagger$ & \textbf{49.44}$^\dagger$ & \underline{23.68} \\
		\rowcolor[gray]{0.9}
		\textbf{Our (soft)}              & \textbf{3.93} & \textbf{44.87}$^\dagger$ & \textbf{0.416}$^\dagger$/\textbf{0.310}$^\dagger$ & \underline{0.022}/\underline{0.087}$^\dagger$ & \textbf{61.17}$^\dagger$ & \multicolumn{1}{c|}{\textbf{19.77}} & \underline{7.13} & \textbf{39.84}$^\dagger$ & \textbf{0.365}$^\dagger$/\textbf{0.253}$^\dagger$  & 0.021/\underline{0.079}$^\dagger$ & \underline{49.27}$^\dagger$ & \underline{23.68} \\ \bottomrule
	\end{tabular}}
	\caption{Results compared across advanced models. \textbf{Bold} text indicates the best performance, and \underline{underlined} text indicates the second-best performance. Significant improvements are marked with $^\dagger$ (t-test, $p < 0.01$) and $^\ddagger$ (t-test, $p < 0.05$). The advanced models labeled $^\diamond$ denote the results from our reproduction.
	}
	\label{mainresult_plm}
\end{table*}

\section{Experiments}
\subsection{Experimental Setting}
\paragraph{Datasets}
In our setup, we require the dataset to be proactively guided and to incorporate elements such as user profiles and domain knowledge. After careful examination, we identify the \textbf{DuRecDial} \cite{liu-etal-2020-towards-conversational} and \textbf{DuRecDial2.0} \cite{liu-etal-2021-durecdial} datasets as suitable benchmarks for our experiments. To further investigate the performance of target-guided dialogue systems, the test set is additionally divided into two subsets: \textbf{In-Domain (ID)} and \textbf{Out-of-Domain (OOD)}. The statistics are summarized in Table~\ref{dataset_two}. For additional details on data processing, please refer to App.~\ref{appedix_dataset}.
\paragraph{Implementation Details}
Please refer to App.~\ref{imple_details}.
\paragraph{Baselines and Metrics}
We compare our framework with advanced models relevant to our task, namely \textbf{TPDial} \cite{wang2023target} and \textbf{TRIPDial} \cite{wang2024target}. We also compare our framework with large language models (LLMs), including LoRA fine-tuning \cite{hulora} (\textbf{LLaMA-1B}, \textbf{LLaMA-3B}, and \textbf{Qwen-3B}) and prompt-based methods (\textbf{LLaMA-8B}, \textbf{Qwen-14B}, and \textbf{Qwen-32B}) \footnote{All models are obtained from the Hugging Face repository.}. Details of the prompts used in this paper are provided in App.~\ref{appendix_prompt}. We adopt \textbf{T5} \cite{raffel2020exploring} (Chinese: \textbf{T5-Zh} \cite{zhao-etal-2023-tencentpretrain}; English: \textbf{T5-Flan} \cite{chung2024scaling}) as our backbone model due to its lightweight design and strong empirical performance.

Following prior work \cite{wang2024target}, we adopt the following widely used automatic evaluation metrics: \textbf{Perplexity (PPL)}; \textbf{Word F1 (W. F1)} \cite{wang-etal-2023-dialogue}; \textbf{BLEU (BLEU-1/2)} \cite{papineni-etal-2002-bleu}; \textbf{Distinct (DIST-1/2)} \cite{li-etal-2016-diversity}; \textbf{Knowledge F1 (K. F1)} \cite{liu-etal-2020-towards-conversational}; and \textbf{Failure (Fail.)} \cite{liu-etal-2021-durecdial}. See App.~\ref{appedix_baseline} and App.~\ref{appedix_metrics} for more details.

\subsection{Results and Analysis}
Table~\ref{mainresult_plm} presents the results compared with advanced models, and details of the values of $m$ and $\delta$ are provided in \S\ref{analysis_of_parameters}. Our framework achieves highly competitive performance on both the DuRecDial and DuRecDial2.0 datasets, demonstrating clear advantages on metrics measuring dialogue quality and informativeness. Specifically, on the ID and OOD test sets of both datasets, our framework (in both soft and hard modes) significantly outperforms strong baseline models, including TPDial and TRIPDial, on W. F1 and BLEU-1/2 metrics, while achieving competitive PPL results, indicating improved dialogue fluency. Furthermore, on the K. F1 metric, which measures knowledge utilization, our framework shows substantial improvement over the backbone model (e.g., on the DuRecDial2.0 ID test set, the soft mode achieves 61.17, while the base T5-Flan achieves 55.26). This demonstrates that by jointly modeling user profiles and domain knowledge to introduce dynamic scenario biases, the system generates more informative utterances. Our framework also exhibits superior robustness in target achievement (as reflected by the Failure metric), highlighting its proactivity and generalization ability. Advanced baseline models (e.g., TPDial and TRIPDial), while performing reasonably on the ID test set, show a dramatic increase in failure rate on the OOD test set (e.g., TPDial's failure rate reaches 99.69\% on the DuRecDial2.0 OOD test set). In contrast, our framework maintains competitive target achievement on OOD data, with failure rates of 20.20\% and 23.68\%, primarily attributed to the IKB, which predicts intent keywords for future dialogue turns. The soft mode, which better accommodates prediction uncertainty, outperforms the hard mode on most core metrics. These results further validate that higher-level, flexible dynamic guidance is crucial for target-guided proactive dialogue systems.

We also conduct comparative experiments with several recent LLMs using both fine-tuning and prompt-based approaches, as summarized in Table~\ref{mainresult_llm}. Our lightweight 0.3B framework demonstrates strong competitiveness against LLMs ranging from 1B to 32B parameters, indicating its capability to perform on par with much larger models. Ultra-large prompt-based models (e.g., Qwen$_{\text{32B}}$), which generate utterances via prompting, often struggle to preserve lexical overlap and dialogue naturalness, leading to substantially lower W. F1 and BLEU scores compared to fine-tuned models. In contrast, our framework excels at maintaining high-quality dialogue flow. Notably, on the ID test set of the DuRecDial2.0 dataset, our soft mode achieves state-of-the-art W. F1 (44.87) and BLEU-1/2 (0.416/0.310), outperforming even parameter-efficient fine-tuned LLaMA$_{\text{3B}}$ and Qwen$_{\text{3B}}$ models. Moreover, it sustains highly competitive knowledge utilization (K. F1), demonstrating that our framework effectively integrates domain knowledge. Our framework exhibits a superior trade-off between dialogue quality and target achievement. On the DuRecDial2.0 dataset, fine-tuned LLMs show high target failure rates (e.g., LLaMA$_{\text{3B}}$ reaches 40.30\% on the ID test set), indicating that fine-tuning alone is insufficient for mastering complex proactive planning. In contrast, large prompt-based models (e.g., Qwen$_{\text{32B}}$) achieve extremely low failure rates (3.80\%) but often at the cost of dialogue quality, abruptly forcing target achievement, as reflected by a sharp drop in dialogue-related metrics. By leveraging multi-turn dynamic guidance, our framework achieves a low failure rate of 19.77\% on the ID test set while maintaining high dialogue coherence. This demonstrates that our framework is not merely a computationally constrained compromise, but effectively balances precise target orientation with natural, human-like interaction, maintaining unique architectural advantages even compared with large-scale LLMs.



\begin{table*}[!t]
	\small
	\centering
	\setlength{\tabcolsep}{0.1mm}{
	\begin{tabular}{lcccccccccccc}
		\toprule
		\multicolumn{13}{c}{\textbf{DuRecDial Dataset}}                                                                                                                            \\ \cmidrule(l){2-13} 
		\multirow{2}{*}{\textbf{Model}} & \multicolumn{6}{c|}{\textbf{ID Test Set}}                                              & \multicolumn{6}{c}{\textbf{OOD Test Set}}                         \\ \cmidrule(l){2-13} 
		& \textbf{PPL$_\downarrow$}  & \textbf{W. F1}  & \textbf{BLEU-1/2}       & \textbf{DIST-1/2}       & \textbf{K. F1}  & \multicolumn{1}{c|}{\textbf{Fail.$_\downarrow$}}  & \textbf{PPL$_\downarrow$}  & \textbf{W. F1}  & \textbf{BLEU-1/2}       & \textbf{DIST-1/2}       & \textbf{K. F1}  & \textbf{Fail.$_\downarrow$}  \\ \midrule
		\textbf{LLaMA$_{\text{1B}}^\diamondsuit$}                & 3.61 & 44.95 & 0.402/0.330 & 0.014/0.104 & 57.24 & \multicolumn{1}{c|}{37.52} & 4.58 & 44.89 & 0.408/0.340 & 0.017/0.107 & 57.25 & 38.90 \\
		\textbf{LLaMA$_{\text{3B}}^\diamondsuit$}                & \textbf{3.32} & 48.09 & 0.434/0.363 & 0.015/0.118 & 63.44 & \multicolumn{1}{c|}{26.29} & \textbf{4.23} & 46.68 & 0.430/0.362 & 0.019/0.124 & 60.96 & 34.41 \\
		\textbf{Qwen$_{\text{3B}}^\diamondsuit$}                 & 3.66 & \textbf{48.99} & \textbf{0.442}/\textbf{0.368} & \textbf{0.016}/\textbf{0.134} & \textbf{64.56} & \multicolumn{1}{c|}{22.01} & 4.57 & \textbf{47.29} & \textbf{0.439}/\textbf{0.367} & \textbf{0.021}/\textbf{0.147} & \textbf{62.93} & 27.43 \\
		\textbf{LLaMA$_{\text{8B}}^\heartsuit$}                & 1.75 & 28.61 & 0.224/0.156 & 0.009/\underline{0.097} & 38.84 & \multicolumn{1}{c|}{15.51} & 1.76 & 27.93 & 0.224/0.156 & \underline{0.012}/\underline{0.106} & 36.78 & 22.69 \\
		\textbf{Qwen$_{\text{14B}}^\heartsuit$}                & 1.53 & 30.17 & 0.221/0.150 & 0.007/0.079 & 45.32 & \multicolumn{1}{c|}{2.81}  & 1.52 & 29.06 & 0.210/0.143 & 0.009/0.089 & 43.29 & 1.75  \\
		\textbf{Qwen$_{\text{32B}}^\heartsuit$}                & \underline{1.46} & 31.20 & 0.230/0.160 & 0.007/0.073 & 42.09 & \multicolumn{1}{c|}{\underline{2.22}}  & \underline{1.45} & 30.36 & 0.222/0.154 & 0.009/0.080 & 40.50 & \underline{1.00}  \\
		\rowcolor[gray]{0.9}
		\textbf{Our$_{\text{0.3B}}$ (hard)}             & 3.89 & 44.70 & \underline{0.433}$^\dagger$/0.347 & \underline{0.010}$^\dagger$/0.061 & 49.62 & \multicolumn{1}{c|}{\textbf{17.13}$^\dagger$} & 5.36 & 44.20 & 0.431/0.347 & \underline{0.012}$^\dagger$/0.066 & \underline{47.83}$^\dagger$ & \textbf{19.70}$^\dagger$ \\
		\rowcolor[gray]{0.9}
		\textbf{Our$_{\text{0.3B}}$ (soft)}             & 3.87 & \underline{44.87}$^\dagger$ & \underline{0.433}$^\dagger$/\underline{0.348}$^\dagger$ & \underline{0.010}$^\dagger$/0.061 & \underline{49.83}$^\dagger$ & \multicolumn{1}{c|}{\textbf{17.13}$^\dagger$} & 5.34 & \underline{44.33}$^\dagger$ & \underline{0.432}$^\dagger$/\underline{0.348}$^\dagger$ & \underline{0.012}$^\ddagger$/0.065 & \underline{47.83}$^\dagger$ & 20.20 \\ \midrule
		\multicolumn{13}{c}{\textbf{DuRecDial2.0 Dataset}}                                                                                                                         \\ \midrule
		\textbf{LLaMA$_{\text{1B}}^\diamondsuit$}                & 3.80 & 42.03 & 0.382/0.282 & 0.031/0.128 & 61.31 & \multicolumn{1}{c|}{49.81} & 4.83 & 39.34 & 0.358/0.256 & \textbf{0.031}/0.129 & 55.19 & 41.12 \\
		\textbf{LLaMA$_{\text{3B}}^\diamondsuit$}                & \textbf{3.44} & 44.32 & 0.397/0.300 & 0.032/0.127 & \textbf{65.60} & \multicolumn{1}{c|}{40.30} & 4.32 & \textbf{40.94} & 0.362/\textbf{0.262} & \textbf{0.031}/0.123 & \textbf{59.57} & 36.14 \\
		\textbf{Qwen$_{\text{3B}}^\diamondsuit$}                 & 3.47 & 43.07 & 0.384/0.283 & \textbf{0.033}/\textbf{0.139} & 63.77 & \multicolumn{1}{c|}{33.65} & \textbf{4.17} & 39.84 & 0.357/0.255 & \textbf{0.031}/\textbf{0.130} & 58.66 & 28.35 \\
		\textbf{LLaMA$_{\text{8B}}^\heartsuit$}                & 1.51 & 26.19 & 0.199/0.109 & 0.019/0.114 & 35.99 & \multicolumn{1}{c|}{3.42}  & 1.53 & 26.03 & 0.196/0.101 & 0.021/0.116 & 32.93 & 5.61  \\
		\textbf{Qwen$_{\text{14B}}^\heartsuit$}                & 1.33 & 26.51 & 0.191/0.097 & 0.021/\underline{0.125} & 42.66 & \multicolumn{1}{c|}{\underline{2.28}}  & 1.34 & 26.38 & 0.192/0.096 & \underline{0.023}/\underline{0.128} & 39.08 & 6.23  \\
		\textbf{Qwen$_{\text{32B}}^\heartsuit$}                & \underline{1.28} & 26.47 & 0.192/0.099 & 0.020/0.113 & 37.19 & \multicolumn{1}{c|}{3.80}  & \underline{1.30} & 25.92 & 0.186/0.093 & 0.020/0.114 & 33.97 & \underline{5.30}  \\
		\rowcolor[gray]{0.9}
		\textbf{Our$_{\text{0.3B}}$ (hard)}             & 3.96 & 44.79 & \textbf{\underline{0.416}$^\dagger$}/0.309 & \underline{0.022}$^\dagger$/0.088 & 61.11 & \multicolumn{1}{c|}{\textbf{19.77}$^\dagger$} & 7.27 & 39.61 & 0.363/0.252 & 0.022/0.082 & \underline{49.44}$^\dagger$ & \textbf{23.68}$^\ddagger$ \\
		\rowcolor[gray]{0.9}
		\textbf{Our$_{\text{0.3B}}$ (soft)}             & 3.93 & \textbf{\underline{44.87}}$^\ddagger$ & \textbf{\underline{0.416}}$^\dagger$/\textbf{\underline{0.310}}$^\dagger$ & \underline{0.022}$^\dagger$/0.087 & \underline{61.17}$^\dagger$ & \multicolumn{1}{c|}{\textbf{19.77}$^\dagger$} & 7.13 & \underline{39.84}$^\dagger$ & \textbf{\underline{0.365}}$^\dagger$/\underline{0.253}$^\dagger$ & 0.021/0.079 & 49.27 & \textbf{23.68}$^\ddagger$ \\ \bottomrule
	\end{tabular}}
	\caption{Results compared with LLMs. Models labeled $^\diamondsuit$ denote fine-tuning, while models labeled $^\heartsuit$ denote the prompt-based method. \textbf{Bold} text indicates the best performance among fine-tuning methods, and \underline{underlined} text indicates the best performance among prompt-based methods. Significant improvements are marked with $^\dagger$ (t-test, $p < 0.01$) and $^\ddagger$ (t-test, $p < 0.05$). 
	}
	\label{mainresult_llm}
\end{table*}

\begin{table*}[!t]
	\small
	\centering
	\setlength{\tabcolsep}{0.9mm}{
		\begin{tabular}{llllllll}
			\toprule
			\textbf{Split}                & \textbf{Model}  & \textbf{PPL$_\downarrow$}       & \textbf{W. F1} & \textbf{BLEU-1/2}    & \textbf{DIST-1/2}    & \textbf{K. F1} & \textbf{Failure$_\downarrow$} \\ \midrule
			\multirow{5}{*}{\textbf{ID Test Set}}  & \textbf{Ours}          & 3.89 & 44.70     & 0.433/0.347 & 0.010/0.061 & 49.62          & 17.13   \\
			& \textbf{w/o IKB}      & 3.32$_{\downarrow0.57}$    & 45.16$_{\uparrow0.46}$      & 0.414/0.328$_{\downarrow0.019/0.019}$ & 0.010/0.050$_{\downarrow0.000/0.011}$ & 52.45$_{\uparrow2.83}$          & 24.52$_{\uparrow7.39}$   \\
			
			& \textbf{w/o CSM} & 3.64$_{\downarrow0.25}$ & 44.67$_{\downarrow0.03}$      & 0.429/0.343$_{\downarrow0.004/0.004}$ & 0.009/0.050$_{\downarrow0.001/0.011}$ & 46.52$_{\downarrow3.10}$          & 16.69$_{\downarrow0.44}$   \\
			& \textbf{-$\mathcal{F}_k(\mathbf{H}^k)$}   & 3.86$_{\downarrow0.03}$  & 44.04$_{\downarrow0.66}$    & 0.427/0.341$_{\downarrow0.006/0.006}$ & 0.010/0.059$_{\downarrow0.000/0.002}$ & 47.40$_{\downarrow2.22}$          & 16.25$_{\downarrow0.88}$   \\
			& \textbf{-$\mathcal{F}_u(\mathbf{H}^u)$}     & 3.56$_{\downarrow0.33}$   & 44.56$_{\downarrow0.14}$     & 0.424/0.337$_{\downarrow0.009/0.010}$ & 0.009/0.047$_{\downarrow0.001/0.014}$ & 48.83$_{\downarrow0.79}$          & 18.32$_{\uparrow1.19}$   \\
			\midrule
			\multirow{5}{*}{\textbf{OOD Test Set}} & \textbf{Ours}    & 5.36      & 44.20      & 0.431/0.347 & 0.012/0.066 & 47.83          & 19.70   \\
			& \textbf{w/o IKB}  & 4.49$_{\downarrow0.87}$        & 43.46$_{\downarrow0.74}$      & 0.411/0.327$_{\downarrow0.020/0.020}$ & 0.011/0.044$_{\downarrow0.001/0.022}$ & 47.87$_{\uparrow0.04}$          & 26.43$_{\uparrow6.73}$   \\
			& \textbf{w/o CSM}  & 4.73$_{\downarrow0.63}$  & 43.82$_{\downarrow0.38}$      & 0.424/0.338$_{\downarrow0.007/0.009}$ & 0.010/0.050$_{\downarrow0.002/0.016}$ & 43.03$_{\downarrow4.80}$          & 20.20$_{\uparrow0.50}$   \\
			& \textbf{-$\mathcal{F}_k(\mathbf{H}^k)$}  & 5.25$_{\downarrow0.11}$    & 43.09$_{\downarrow1.11}$     & 0.421/0.335$_{\downarrow0.010/0.012}$ & 0.012/0.065$_{\downarrow0.000/0.001}$ & 43.95$_{\downarrow3.88}$          & 17.96$_{\downarrow1.74}$   \\
			& \textbf{-$\mathcal{F}_u(\mathbf{H}^u)$}  & 4.64$_{\downarrow0.72}$    & 44.06$_{\downarrow0.14}$     & 0.424/0.339$_{\downarrow0.007/0.008}$ & 0.010/0.044$_{\downarrow0.002/0.022}$ & 47.67$_{\downarrow0.16}$          & 19.70$_{\downarrow0.00}$   \\
			\bottomrule
	\end{tabular}}
	\caption{Ablation study on the DuRecDial dataset.}
	\label{ablation_durecdial}
\end{table*}

\begin{figure}[!t]
	\centering
	\includegraphics[width=\columnwidth]{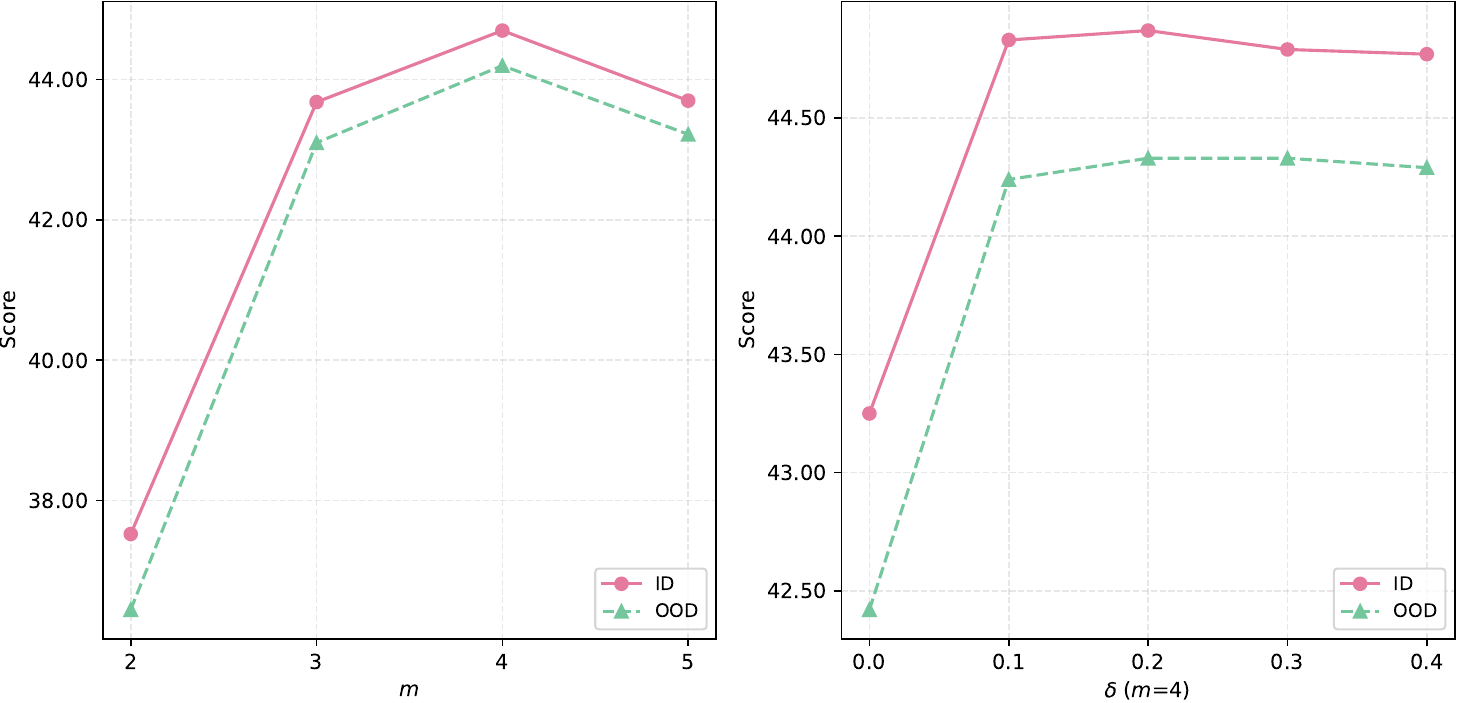}
	\caption{
		Variation of W. F1 w.r.t $m$ and $\delta$.
	}
	\label{hy_durecdial}
\end{figure}

\subsection{Analysis of Parameters}
\label{analysis_of_parameters}


We analyze two key hyperparameters, $m$ and $\delta$. Specifically, $m$ specifies the number of future turns for which the framework predicts intent keywords, while $\delta$ defines the probability threshold for selecting intent keywords during inference in the soft mode, where only intent keywords with probabilities exceeding $\delta$ are retained, as detailed in \S\ref{hard_soft_modes}. We adopt W. F1 as the primary performance metric, as it captures exact word-level overlap. The performance trends measured by W. F1 on the DuRecDial and DuRecDial2.0 datasets, as $m$ and $\delta$ vary, are presented in Figures~\ref{hy_durecdial} and~\ref{hy_durecdial2.0}, respectively. We choose $\delta$ in the range of 0.0–0.4, as values above 0.5 have a high probability of being selected according to the sigmoid principle. Figure~\ref{hy_durecdial} shows that our framework remains stable across both ID and OOD test sets of the DuRecDial dataset. Performance peaks at $m=4$: an excessively small $m$ fails to provide sufficient forward-looking guidance, while an overly large $m$ (e.g., $m=5$) introduces noise from distant future turns. Using $m=4$, we further examine performance variations with respect to $\delta$, resulting in a selected value of $\delta=0.2$, which effectively filters low-confidence noise without sacrificing informative keywords. Similarly, Figure~\ref{hy_durecdial2.0} presents performance variations on the DuRecDial2.0 dataset, where the framework achieves the best balance of guidance at $m=3$. Accordingly, we set $m=3$ and analyze performance changes with respect to $\delta$. Considering the overall results on both ID and OOD test sets, we set $\delta=0.2$. We also conduct more comprehensive experiments and analyses, as reported in App.~\ref{full_experiments}.

\subsection{Ablation Study}
\label{ablation_study_main}
We conduct ablation studies to verify the effects of the components proposed in this paper. Specifically, we investigate the following variants: (1) without IKB (w/o IKB); (2) without CSM (w/o CSM); (3) without domain knowledge modeling ($-\mathcal{F}_k(\mathbf{H}^k)$); and (4) without user profile modeling ($-\mathcal{F}_u(\mathbf{H}^u)$). The results on the DuRecDial and DuRecDial2.0 datasets, reported in Tables~\ref{ablation_durecdial} and~\ref{ablation_durecdial2.0}, respectively, clearly demonstrate the core utility of CSM and IKB in improving system utterance informativeness, fluency, and proactivity. For informativeness and fluency, removing CSM (w/o CSM) significantly degrades the framework's knowledge utilization (K. F1) on both the ID and OOD test sets across the two datasets. For example, on the DuRecDial2.0 dataset, K. F1 drops sharply by 7.71 and 6.97, respectively. Fine-grained analysis further shows that removing domain knowledge modeling ($-\mathcal{F}_k(\mathbf{H}^k)$) directly leads to notable decreases in K. F1 and BLEU scores, while removing user profile modeling ($-\mathcal{F}_u(\mathbf{H}^u)$) weakens lexical diversity (DIST-1/2) and increases the target failure rate. These results indicate that CSM effectively integrates contextual information through dynamic scenario bias, thereby avoiding the homogenization of utterances. For proactivity, IKB serves as a key guiding engine. Removing IKB leads to a substantial increase in the target failure rate on both the DuRecDial and DuRecDial2.0 datasets, with increases of 7.39 and 6.73 on the ID and OOD test sets of the DuRecDial dataset, respectively. Moreover, the absence of IKB also degrades dialogue quality metrics such as W. F1 and BLEU. This suggests that IKB does not mechanically enforce topic transitions; instead, it enables smooth and coherent proactive guidance through flexible multi-step intent keyword predictions, thereby narrowing the gap between target-guided proactive dialogue systems and natural real-world interactions.

\subsection{Human Evaluation}

\begin{figure}[!t]
	\small
	\centering
	\includegraphics[width=\linewidth]{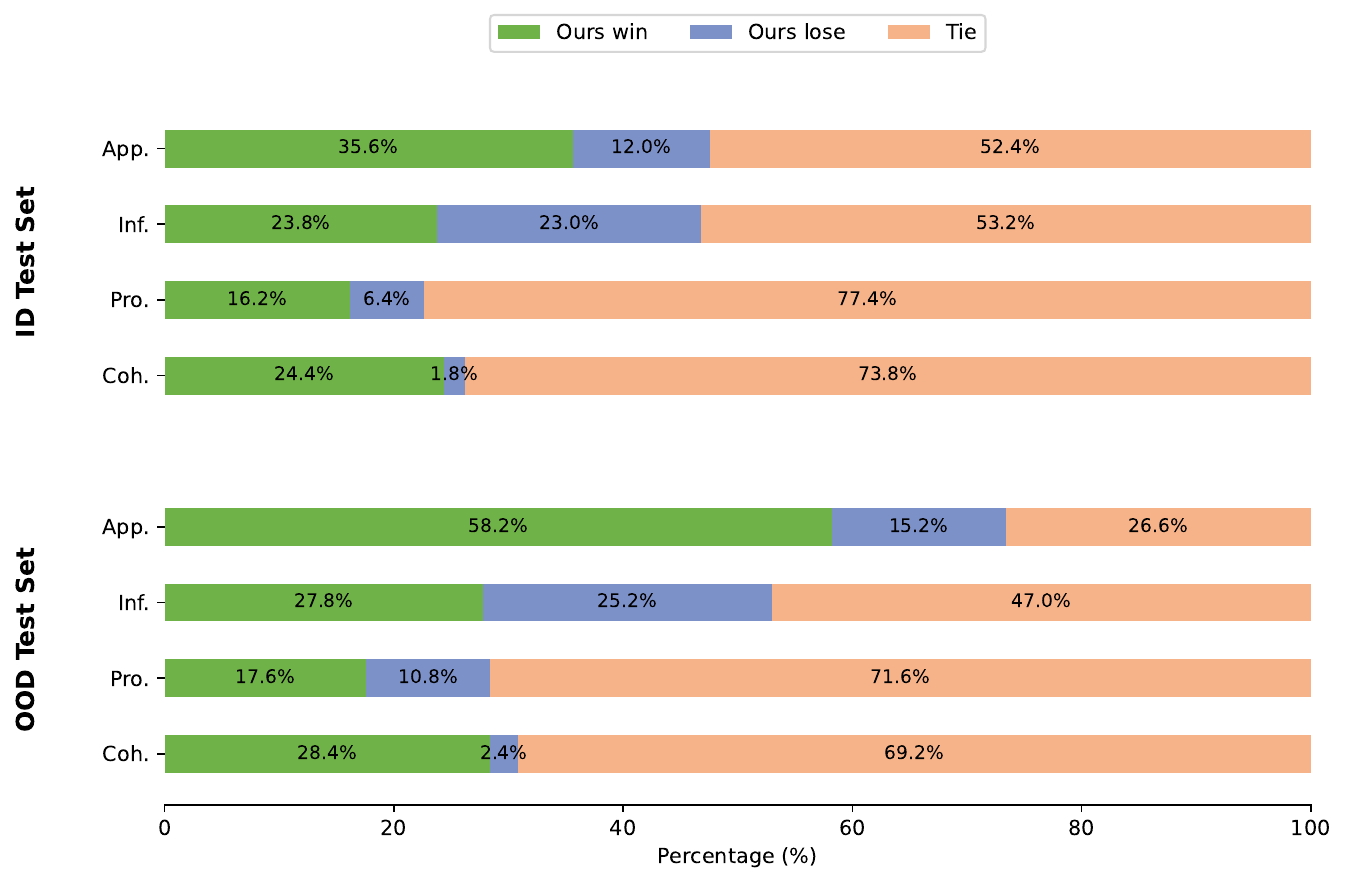}
	\caption{
		Human evaluation. "App." denotes "appropriateness," "Inf." denotes "informativeness," "Pro." denotes "proactivity," and "Coh." denotes "coherence." 
	}
	\label{human_evaluate_durecdial2.0}
\end{figure}

We perform pairwise human evaluation to further validate our framework. Specifically, we compare our framework with the baseline T5-Flan. "Win," "Lose," and "Tie" indicate that our framework outperforms T5-Flan, underperforms relative to T5-Flan, and performs comparably to T5-Flan, respectively. Figure~\ref{human_evaluate_durecdial2.0} presents the human evaluation results. In terms of coherence, which measures dialogue fluency, our framework achieves win rates of 24.4\% and 28.4\% on the ID and OOD test sets, respectively, substantially outperforming the baseline model's 1.8\% and 2.4\%. For appropriateness and informativeness, which reflect how well the content fits the dialogue context (intent keywords, domain knowledge, and user profiles), our framework maintains its advantage on the ID test set and achieves an impressive win rate of 58.2\% in appropriateness on the OOD test set (baseline: 15.2\%). These results suggest that CSM effectively leverages domain knowledge and user profiles to enrich utterance content under unseen data distributions. In terms of proactivity, our framework achieves clear net wins over the baseline on both the ID (16.2\% vs. 6.4\%) and OOD (17.6\% vs. 10.8\%) test sets. Some ties occur due to the strong foundation of the base skeleton model; however, by leveraging the higher-level multi-step intent keyword guidance provided by IKB, our framework reduces mechanical or abrupt topic shifts, thereby narrowing the gap between target-guided proactive dialogue systems and natural interactions. Details of the human evaluation procedure are provided in App.~\ref{human_evaluate_appendix}.

\begin{figure*}[!t]
	\centering
	\includegraphics[width=\textwidth]{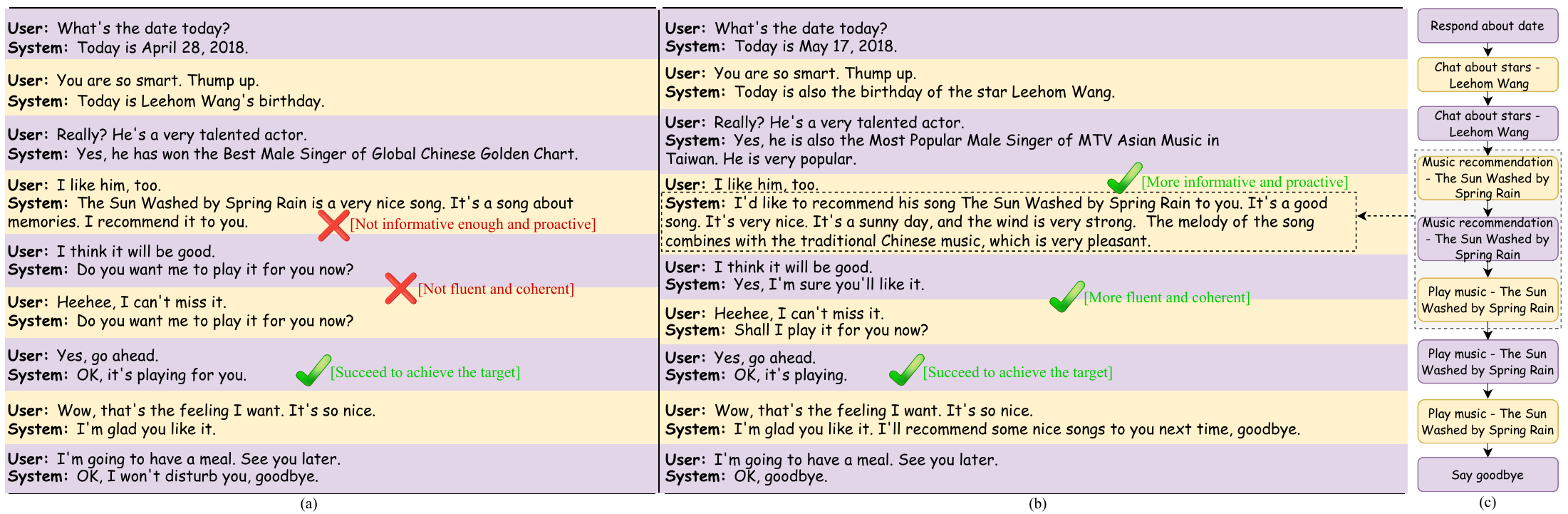}
	\caption{
		Case study (ID test set). (a) T5-Flan; (b) Our framework; (c) Intent keyword transitions.
	}
	\label{case1}
\end{figure*}

\subsection{Case Study}
\label{main_case_study}
Figure~\ref{case1} presents an example showing how our framework outperforms the baseline T5-Flan in real-world dialogues. When recommending songs, T5-Flan's utterances lack detail and awkwardly repeat the question \textit{Do you want me to play it for you now?} in subsequent turns, revealing interaction rigidity. In contrast, CSM effectively integrates underlying knowledge. As the ablation results indicate that removing CSM leads to a sharp drop in knowledge utilization (K. F1), our framework generates more detailed utterances such as \textit{It's a sunny day} and \textit{combines with the traditional Chinese music}, enhancing informativeness and user engagement. IKB provides the system with a flexible, multi-step forward-looking perspective, allowing the framework to naturally connect with the user's expectations in the fifth turn, producing \textit{Yes, I'm sure you'll like it} before logically proposing a playback request. This guidance trajectory is smooth, avoiding abrupt topic shifts and maintaining dialogue quality. The case study demonstrates the effectiveness of our framework, narrowing the gap between target-guided dialogue systems and natural real-world interactions. A more detailed analysis is provided in App.~\ref{case_study}.

\section{Related Work}
We focus on the proactivity and target-guided aspects of dialogue; thus, we review related work on proactive and target-guided dialogue systems.

Proactive dialogue differs from traditional dialogue systems that merely respond passively to users; it requires the system to actively steer the conversation toward accomplishing specific tasks, emphasizing initiative \cite{tang-etal-2025-kapa,deng2025proactive}. One line of research in proactive dialogue focuses on developing various types of dialogue systems, including target-guided dialogue \cite{zhang-etal-2025-enhancing-goal}, prosocial dialogue \cite{ziems-etal-2022-moral}, non-collaborative dialogue \cite{jin-etal-2024-persuading}, and user preference elicitation \cite{zhang2024towards}. These systems primarily leverage domain knowledge or user profiles to enhance proactivity \cite{liu-etal-2021-durecdial}, or employ planning-based strategies \cite{wang2024target}. In this work, we enhance the system's proactivity by modeling conversational scenarios and employing intent–keyword bridging for dynamic guidance.


Target-guided dialogue requires the system to generate utterances that fulfill a pre-determined target, such as incorporating a keyword into the system utterance. Accordingly, a keyword is typically designated as the target, making keyword planning the natural primary strategy. To this end, entity-based keywords have been widely adopted, and a variety of planning strategies have been proposed \cite{tang-etal-2019-target,zhong2021keyword}. \citet{zhong2021keyword} further leveraged ConceptNet \cite{speer2017conceptnet} to support next-turn keyword planning, while both local and global keyword planning approaches have demonstrated substantial progress \cite{yang-etal-2022-topkg}. However, the semantic information conveyed by entity-based keywords is limited, motivating the adoption of keywords with richer and more abstract semantics \cite{dao-etal-2023-reinforced,wang2024target,zhang-etal-2025-enhancing-goal}. In this way, planning sequences of such abstract keywords has become a primary research focus. For example, \citet{wang-etal-2023-dialogue} proposed Brown bridge planning, \citet{wang2023target} introduced target-driven planning, \citet{dao-etal-2023-reinforced} developed global reinforcement learning, \citet{zhang2024goal} presented graph-interaction planning, and \citet{wang2024target,kang2026pseudo} proposed bidirectional planning. However, the resulting keyword sequences remain static, relying only on one-turn keywords without looking ahead, leading to a notable mismatch with the dynamics of real-world conversations. To address this, we jointly model user profiles and domain knowledge as conversational scenarios, and leverage dynamic intent–keyword bridging to further reduce this gap.

\section{Conclusion}
In this work, we jointly model user profiles and domain knowledge as conversational scenarios, which introduce a bias to dynamically influence utterances, and we propose intent-keyword bridging to predict intent keywords for upcoming turns, providing higher-level and flexible guidance. Extensive experiments show that our framework substantially enhances target-guided proactive dialogue systems, alleviating the gap with real-world interactions.

\section*{Limitations}
We validate the effectiveness of our framework through extensive automatic and human evaluations; nevertheless, several limitations remain. The datasets used in this work are built upon \cite{wang2023target,wang2024target}, from which we introduce the concept of intent keywords. We argue that these intent keywords capture higher-level abstract semantics, providing more effective guidance. However, we do not further categorize the intent keywords in the datasets following \cite{wang2024target}, which constitutes a potential limitation of this study. In addition, while our framework does not consistently outperform LLM-based models, it integrates the strengths of fine-tuned LLMs and prompt-based LLMs. For instance, in terms of dialogue quality and failure rate, it achieves competitive results with only 0.3B parameters. Although its contribution to advancing state-of-the-art performance is limited, it remains valuable for research in low-resource settings. The experiments collectively demonstrate the effectiveness of our proposed mechanisms, particularly CSM and IKB, which are of clear significance. In summary, potential directions for improvement include: (1) further categorizing and refining intent keywords while exploring the adoption of a unified and powerful reasoning mechanism under unpredictable user behavior; (2) adopting more specialized extraction methods for conversational scenario modeling, where bias introduction may require a more "soft" approach; and (3) exploring LLM-based backbone models, which are not employed in this work due to resource constraints. We believe that addressing these issues will further advance research on target-guided proactive dialogue systems.

Furthermore, given the proactive nature of dialogue in guiding users and its potential negative social impacts, such as political interference and abuse, we strongly recommend minimizing such misuse and strictly adhering to relevant laws. This study is limited to academic research.

\bibliography{custom}

\appendix

\section{Datasets}
\label{appedix_dataset}
Both the original \textbf{DuRecDial} \cite{liu-etal-2020-towards-conversational} and \textbf{DuRecDial2.0} \cite{liu-etal-2021-durecdial} datasets were collected via crowdsourced human–human conversations, where one participant acted as the user and the other as the system. In these conversations, the user was provided with profile information such as preferences and age, while the system received domain knowledge consisting of domain-specific topics and their associated attributes (e.g., movies, music). The original DuRecDial dataset contains approximately 10k Chinese dialogues and 156k utterances, whereas DuRecDial2.0 dataset includes 8.2k dialogues in both Chinese and English. Both datasets offer keyword-type and keyword-topic annotations throughout the conversations, making them well suited for the target-guided proactive dialogue task investigated in this work. Incidentally, in this type of dataset, the dialogue target often corresponds to the keyword-topics appearing in the last system utterance, and this is how we determine whether the system has achieved its dialogue target.

To better evaluate target-guided proactive dialogue systems, \citet{wang2023target} repurposed these two datasets by filtering out dialogues that introduced no new keyword-topics, thereby ensuring that each intent keyword was grounded in domain-specific knowledge within the conversation. They further enriched the domain knowledge by sampling additional information from triplets within a two-hop range of the target keyword-topic in the domain knowledge graph. To more thoroughly investigate the performance of target-guided proactive dialogue systems, the test set was additionally divided into two subsets: \textbf{In-Domain (ID)}, where the target keyword-topics in the test set were allowed to appear in the training data, and \textbf{Out-of-Domain (OOD)}, where none of the target keyword-topics in the test set appeared in the training set. After processing, the DuRecDial dataset yielded 13 keyword-types and 640 keyword-topics, while the processed DuRecDial2.0 dataset contained 13 keyword-types and 628 keyword-topics. On average, each dialogue exhibited approximately 4.3–4.8 intent keyword transitions from the beginning to the end of the conversation. For the DuRecDial2.0 dataset, we used only the English version, resulting in two dataset variants for a more comprehensive evaluation. Since \citet{wang2023target,wang2024target} provided a further cleaned and processed version of the two datasets, we conducted our experiments using these processed releases.

\begin{table}[!t]
	\centering
	\setlength{\tabcolsep}{0.4mm}{
		\begin{tabular}{llccc}
			\toprule
			\multicolumn{2}{c}{\textbf{Dataset}}              & \textbf{\#Dial.} & \textbf{\#Utter.} & \textbf{Trans.\#Avg.} \\ \midrule
			\multirow{4}{*}{\textbf{DuRecDial}}    & Train    & 4440    & 72466    & 4.4                        \\
			& Dev      & 633     & 10467    & 4.5                        \\
			& ID  & 780     & 12633    & 4.4                        \\
			& OOD & 486     & 7500     & 4.3                        \\ \midrule
			\multirow{4}{*}{\textbf{DuRecDial2.0}} & Train    & 4256    & 68781    & 4.4                        \\
			& Dev      & 608     & 9677     & 4.3                        \\
			& ID  & 770     & 12299    & 4.3                        \\
			& OOD & 446     & 7962     & 4.8                        \\ \bottomrule
	\end{tabular}}
	\caption{Statistics of the DuRecDial and DuRecDial2.0 datasets. Here, "Dial." denotes "dialogue," "Utter." denotes "utterance," and "Trans." denotes "keyword transition."}
	\label{dataset_two}
\end{table}

\begin{figure}[!t]
	\centering
	\includegraphics[width=\columnwidth]{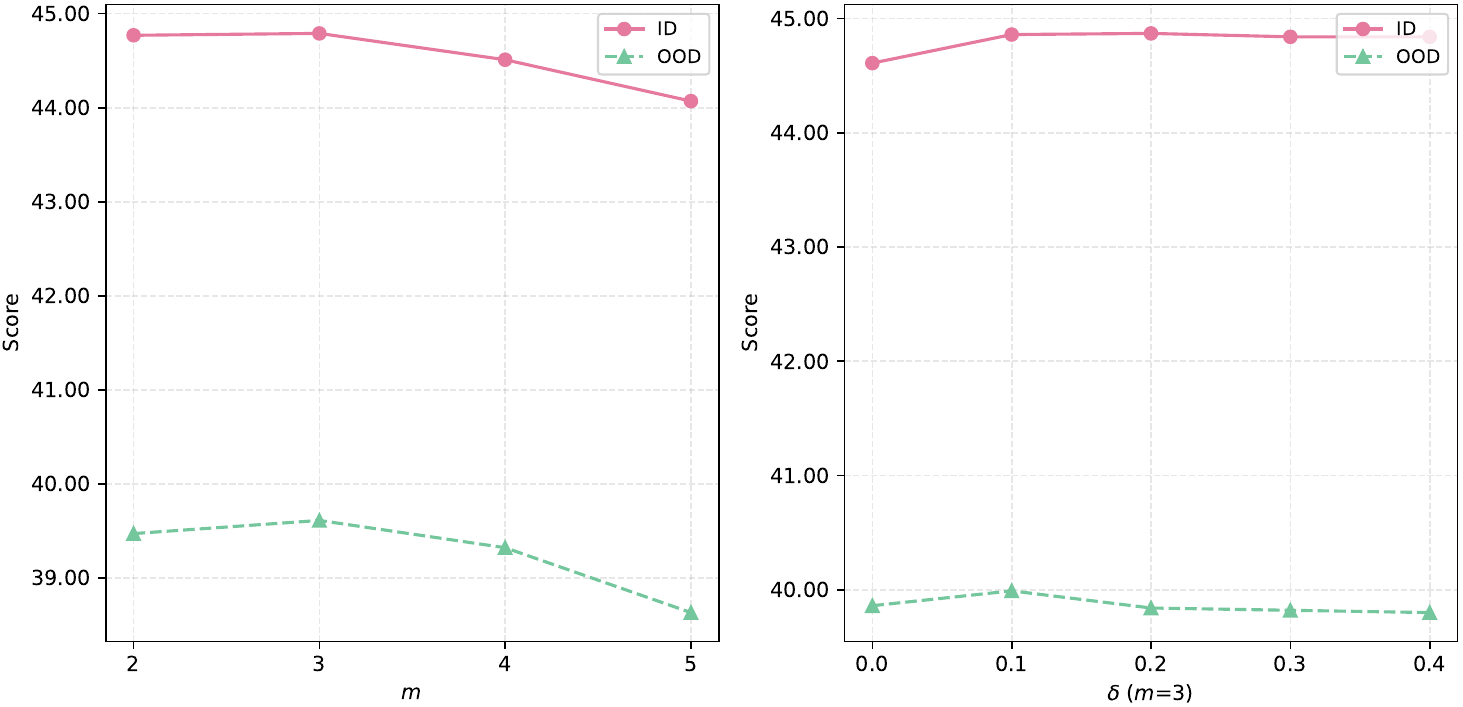}
	\caption{
		Variation of Word F1 with respect to $m$ and $\delta$ on the DuRecDial2.0 dataset.
	}
	\label{hy_durecdial2.0}
\end{figure}

\section{Implementation Details}
\label{imple_details}
We use AdamW as the optimizer and employ both a warmup strategy and gradient clipping. For our framework, training is performed for 50 epochs with a batch size of 8 and a learning rate of $3 \times 10^{-5}$. The hyperparameters $m$ and the threshold $\delta$ are determined empirically; see \S \ref{analysis_of_parameters} for details. We fine-tune LLaMA-1B, LLaMA-3B, and Qwen-3B using the LoRA method \cite{hulora}, with a LoRA rank of 8 and a LoRA alpha of 16. Training is conducted with a batch size of 1, a learning rate of $1 \times 10^{-4}$, and 10 epochs, following common community practice. For the DuRecDial dataset, the maximum decoding length during generation is set to 100, whereas for the DuRecDial2.0 dataset it is set to 80. During model training, all experiments are conducted on a GeForce RTX 3090 GPU. For the prompt-based methods, when GPU memory is insufficient, the model parameters are distributed across a GeForce RTX 3090 GPU and a GeForce RTX 4090 GPU for inference.

Our framework employs different base models as backbones for the Chinese and English datasets. For the DuRecDial dataset, we adopt T5-Zh \cite{zhao-etal-2023-tencentpretrain} as the backbone, since the original T5 \cite{raffel2020exploring} does not support Chinese corpora, whereas the DuRecDial dataset is a linguistically rich and complex Chinese dataset. For the DuRecDial2.0 dataset, we use T5-Flan \cite{chung2024scaling} as the backbone, as instruction tuning enables T5-Flan to substantially outperform T5, despite supporting only English. We select these models as backbones in our experiments because they are strong foundational models in recent years and well-aligned with the design requirements of our framework. The intent keyword hidden states $\mathbf{H}^z$ are concatenated with $\mathbf{K}$ and $\mathbf{V}$ in the cross-attention of the T5 decoder, as illustrated in Figure~\ref{Frame_1}. During training, each backbone model is initialized with its pre-trained parameters and further fine-tuned. For the two state-of-the-art models, TPDial \cite{wang2023target} and TRIPDial \cite{wang2024target}, which are highly relevant to our task, we reproduce them using the code released by their authors, with parameter settings consistent with the original descriptions.


\section{Baselines}
\label{appedix_baseline}
\textbf{TPDial} \cite{wang2023target} is a target-driven keyword planning model for guided system-utterance generation, whereas \textbf{TRIPDial} \cite{wang2024target} is a target-constrained bidirectional planning approach for guided system-utterance generation. Since TPDial and TRIPDial encompass multiple models in their original works, we cite only the relatively better-performing variants. Specifically, for TRIPDial, we reference the model in its controlled mode, while for TPDial, we cite the variant with a GPT-2 backbone. \textbf{LLaMA} family and \textbf{Qwen} family are advanced language models developed in recent years, and their 1B/3B variants—known for strong performance and a favorable trade-off—are adopted as our fine-tuning baseline models. We also adopt larger models from these families, namely LLaMA-8B, Qwen-14B, and Qwen-32B, as prompt-based models; the specific prompts are provided in App.~\ref{appendix_prompt}. This setup considers both efficient fine-tuning (details of which are provided in App.~\ref{imple_details}) and the increasingly popular prompt-based methods, allowing for a more comprehensive comparison. \textbf{T5-Zh} \cite{zhao-etal-2023-tencentpretrain} is a Chinese-adapted variant pretrained on large-scale Chinese corpora, whereas \textbf{T5-Flan} \cite{chung2024scaling} is an instruction-tuned extension of T5 that exhibits strong zero-shot and instruction-following capabilities across diverse tasks. We adopt T5-Zh as the backbone for the DuRecDial dataset and T5-Flan as the backbone for the DuRecDial2.0 dataset.

\section{Evaluation Metrics}
\label{appedix_metrics}
Following prior work \cite{wang2023target,wang2024target} and our formalized setting for evaluating target-guided proactive dialogue generation, we adopt the following widely used automatic evaluation metrics:
\begin{itemize}
	\item \textbf{Perplexity (PPL)} evaluates the language model's uncertainty in predicting the token sequence; lower values indicate better fluency and generation quality;
	\item \textbf{Word F1 (W. F1)} \cite{wang-etal-2023-dialogue} measures the exact word-level overlap between generated and reference utterances;
	\item \textbf{BLEU (BLEU-1/2)} \cite{papineni-etal-2002-bleu} evaluates the n-gram overlap between generated and reference utterances;
	\item \textbf{Distinct (DIST-1/2)} \cite{li-etal-2016-diversity} evaluates the lexical diversity of generated utterances;
	\item \textbf{Knowledge F1 (K. F1)} \cite{liu-etal-2020-towards-conversational} measures the correctness of generated knowledge against the ground-truth domain knowledge;
	\item \textbf{Failure (Fail.)} \cite{liu-etal-2021-durecdial} quantifies the probability that the target is not achieved by the end of the conversation.
\end{itemize}





\section{Analysis of Parameters}
\label{full_experiments}
In the main text, we determine the values of $m$ and $\delta$ based on changes in Word F1. Here, we report all results to provide a more comprehensive analysis.

Tables~\ref{m_full_durecdial} and~\ref{m_full_durecdial2.0} present how the performance on the DuRecDial and DuRecDial2.0 datasets varies with different values of $m$. On the DuRecDial dataset, the majority of the best performance is achieved at $m=4$ for both the ID and OOD test sets, whereas on the DuRecDial2.0 dataset, most of the optimal performance occurs at $m=3$ for both test sets. For the DuRecDial dataset, when $m=2$ or $m=3$, predicting intent keywords for the next two or three steps may not provide sufficiently informative guidance. As $m$ increases, performance generally improves and reaches a peak at $m=4$. When $m=5$, additional noise is introduced by predicting intent keywords too far ahead, resulting in a performance drop; nevertheless, the results remain better than those observed with $m=2$ or $m=3$. It is worth noting that on the DuRecDial dataset, when $m=2$, the performance drops dramatically, falling even below that of the vanilla T5-Zh model. This is because predicting only the next two intent keywords provides very limited guidance and further increases the sparsity of the already sparse embeddings ($\text{Emb}_a(\cdot)$ and $\text{Emb}_t(\cdot)$). This effect is more pronounced on the DuRecDial dataset than on the DuRecDial2.0 dataset. Consequently, this phenomenon is not observed on the DuRecDial2.0 dataset, indicating that our framework exhibits more stable performance on this dataset. For the DuRecDial2.0 dataset, competitive performance is obtained at $m=2$ and peaks at $m=3$. Performance then declines at $m=4$ and $m=5$. These results indicate that the sparsity of embeddings has only a limited effect on the stability of performance on the DuRecDial2.0 dataset. Notably, when $m=2$, although the framework achieves the best lexical diversity, this also suggests that intent-keyword bridging may over-constrain the system's utterance, thereby reducing lexical diversity. When $m=3$, the provided guidance strikes a balance across multiple metrics. In contrast, when $m=4$ or $m=5$, the introduced noise leads to an overall decline in performance.

Tables~\ref{d_full_durecdial} and~\ref{d_full_durecdial2.0} show how the performance on the DuRecDial and DuRecDial2.0 datasets varies with different values of $\delta$ when $m$ is fixed to its optimal setting. On the DuRecDial dataset, most of the best performance is achieved at $\delta=0.2$ for both the ID and OOD test sets, while on the DuRecDial2.0 dataset, the majority of the optimal performance also occurs at $\delta=0.2$ for both test sets. In the soft mode, $\delta$ accounts for the uncertainty of the prediction. In general, a larger $\delta$ yields more concise guidance information at the cost of losing certain details, whereas a smaller $\delta$ offers richer information but inevitably introduces noise. For the DuRecDial dataset, when considering all evaluation metrics, the suboptimal performance is primarily observed at $\delta=0.1$ on the ID test set and at $\delta=0.3$ on the OOD test set. This is because, on the ID test set, keyword–topics are allowed to appear in the training set, so providing more information—even if noisy—still yields positive effects; whereas on the OOD test set, more concise information is typically offer clearer and more direct guidance. For the DuRecDial2.0 dataset, on the ID test set, the suboptimal performance is mostly observed at $\delta=0.1$ across all metrics; whereas on the OOD test set, the pattern differs slightly, with the optimal results mostly appearing at $\delta=0.1$ and the suboptimal results at $\delta=0.2$. The explanation is that the ID test set can benefit from additional information—even if noisy—which may positively influence the results, while the OOD test set requires more concise information to provide more direct guidance. Although the specific values of $\delta$ differ between the DuRecDial and DuRecDial2.0 datasets, the overall trend remains consistent. From the ablation study (\S\ref{ablation_study_main}), we observe that intent–keyword bridging, which connects dialogue history to upcoming utterances, affects multiple aspects of the generated utterances, including dialogue fluency, lexical diversity, user engagement, and target achievement, highlighting the benefits of higher-level and flexible guidance. The magnitude of this effect varies with different values of $m$ and $\delta$.

\begin{table*}[!t]
\centering
\setlength{\tabcolsep}{0.8mm}{
\begin{tabular}{llcccccc}
\toprule
\textbf{Split}                &    & \textbf{PPL}$_\downarrow$  & \textbf{Word F1} & \textbf{BLEU-1/2}       & \textbf{DIST-1/2}       & \textbf{Knowledge F1} & \textbf{Failure$_\downarrow$}  \\ \midrule
\multirow{4}{*}{\textbf{ID Test Set}}  & $m=2$ & 5.14 &  37.52   & 0.355/0.274 & \underline{0.009}/0.059 & 46.75   & 19.05 \\
                     & $m=3$ & 4.03    & 43.68   & 0.424/0.339 & \textbf{0.010}/\textbf{0.065} & \underline{49.44}   & 18.32 \\
                     & $m=4$ &  \underline{3.89}    & \textbf{44.70}   & \textbf{0.433}/\textbf{0.347} & \textbf{0.010}/0.061 & \textbf{49.62}   & \textbf{17.13} \\
                     & $m=5$ & \textbf{3.88}   &   \underline{43.70}   & \underline{0.428}/\underline{0.344} & \underline{0.009}/\underline{0.063} & 48.87   & \underline{18.17} \\ \midrule
\multirow{4}{*}{\textbf{OOD Test Set}} & $m=2$ & 7.39    & 36.44   & 0.350/0.271 & \underline{0.011}/0.061 & 45.77   & 21.45 \\
                     & $m=3$ &  \textbf{5.24}  &  43.10   & 0.425/\underline{0.343} & \textbf{0.012}/\textbf{0.068} & \underline{47.61}   & \underline{19.95} \\
                     & $m=4$ &   5.36  &  \textbf{44.20}   & \textbf{0.431}/\textbf{0.347} & \textbf{0.012}/\underline{0.066} & \textbf{47.83}   & \textbf{19.70} \\
                     & $m=5$ &   \underline{5.34}  & \underline{43.22}   & \underline{0.427}/\underline{0.343} & \textbf{0.012}/\underline{0.066} & 46.90   & 20.95 \\ \bottomrule
\end{tabular}}
\caption{Full experiments on the DuRecDial dataset with different $m$ values. \textbf{Bold} text highlights the best results, while \underline{underlined} text indicates the second-best.}
\label{m_full_durecdial}
\end{table*}

\begin{table*}[!t]
\centering
\setlength{\tabcolsep}{0.8mm}{
\begin{tabular}{llcccccc}
\toprule
\textbf{Split}                & $m=4$    & \textbf{PPL}$_\downarrow$   & \textbf{Word F1} & \textbf{BLEU-1/2}       & \textbf{DIST-1/2}       & \textbf{Knowledge F1} & \textbf{Failure$_\downarrow$}  \\ \midrule
\multirow{5}{*}{\textbf{ID Test Set}}  & $\delta=0.0$ & 3.96   & 43.25   & 0.416/0.329 & \textbf{0.010}/\textbf{0.065} & 48.42   & 28.06 \\
                     & $\delta=0.1$ & \textbf{3.86}   &  \underline{44.83}   & \underline{0.432}/\underline{0.347} & \textbf{0.010}/\underline{0.062} & \underline{49.72}   & \underline{17.43} \\
                     & $\delta=0.2$ &   \underline{3.87}  &   \textbf{44.87}   & \textbf{0.433}/\textbf{0.348} & \textbf{0.010}/0.061 & \textbf{49.83}   & \textbf{17.13} \\
                     & $\delta=0.3$ &  3.88   &   44.79   & \textbf{0.433}/\underline{0.347} & \textbf{0.010}/0.061 & 49.69   & \textbf{17.13} \\
                     & $\delta=0.4$ &    \underline{3.87}  &  44.77   & \underline{0.432}/\underline{0.347} & \textbf{0.010}/0.061 & 49.62   & \textbf{17.13} \\ \midrule
\multirow{5}{*}{\textbf{OOD Test Set}} & $\delta=0.0$ &  5.51  & 42.42   & \underline{0.412}/\underline{0.327} & \textbf{0.013}/\textbf{0.072} & 46.29   & 29.43 \\
                     & $\delta=0.1$ & \underline{5.35}  &  44.24   & \textbf{0.432}/\textbf{0.348} & \underline{0.012}/\underline{0.065} & 47.74   & 20.45 \\
                     & $\delta=0.2$ &  \textbf{5.34}  &   \textbf{44.33}   & \textbf{0.432}/\textbf{0.348} & \underline{0.012}/\underline{0.065} & \underline{47.83}   & 20.20 \\
                     & $\delta=0.3$ &  \underline{5.35} &  \textbf{44.33}   & \textbf{0.432}/\textbf{0.348} & \underline{0.012}/0.064 & \textbf{47.95}   & \underline{20.20} \\
                     & $\delta=0.4$ &  \textbf{5.34}  &   \underline{44.29}   & \textbf{0.432}/\textbf{0.348} & \underline{0.012}/\underline{0.065} & \textbf{47.95}   & \textbf{19.95} \\ \bottomrule
\end{tabular}}
\caption{Full experiments on the DuRecDial dataset with different $\delta$ values (when $m=4$). \textbf{Bold} text highlights the best results, while \underline{underlined} text indicates the second-best.}
\label{d_full_durecdial}
\end{table*}

\begin{table*}[!t]
\centering
\setlength{\tabcolsep}{0.8mm}{
\begin{tabular}{llcccccc}
\toprule
\textbf{Split}                &  & \textbf{PPL}$_\downarrow$   & \textbf{Word F1} & \textbf{BLEU-1/2}       & \textbf{DIST-1/2}       & \textbf{Knowledge F1} & \textbf{Failure$_\downarrow$}  \\ \midrule
\multirow{4}{*}{\textbf{ID Test Set}}  & $m=2$ & 4.08 & \underline{44.77}   & \underline{0.414}/\underline{0.308} & \textbf{0.023}/\textbf{0.092} & 57.62   & 23.00 \\
                     & $m=3$   & \underline{3.96}   & \textbf{44.79}   & \textbf{0.416}/\textbf{0.309} & \underline{0.022}/\underline{0.088} & \textbf{61.11}   & \textbf{19.77} \\
                     & $m=4$ &   \textbf{3.94} & 44.51   & 0.412/0.306 & 0.020/0.077 & \underline{60.71}   & 20.72 \\
                     & $m=5$ &  3.99  & 44.07   & 0.408/0.302 & 0.019/0.074 & 59.98   & \underline{20.15} \\ \midrule
\multirow{4}{*}{\textbf{OOD Test Set}} & $m=2$ &  7.62   & 39.47   & \underline{0.362}/\underline{0.251} & \textbf{0.023}/\textbf{0.085} & 46.56   & 29.28 \\
                     & $m=3$ &  7.27  &  \textbf{39.61}   & \textbf{0.363}/\textbf{0.252} & \underline{0.022}/\underline{0.082} & \textbf{49.44}   & \textbf{23.68} \\
                     & $m=4$ &   \textbf{6.80}  &  \underline{39.32}   & 0.360/0.250 & 0.018/0.066 & \underline{47.95}   & 26.48 \\
                     & $m=5$ &   \underline{6.91} &  38.63   & 0.353/0.243 & 0.018/0.066 & 46.71   & \underline{25.23} \\ \bottomrule
\end{tabular}}
\caption{Full experiments on the DuRecDial2.0 dataset with different $m$ values. \textbf{Bold} text highlights the best results, while \underline{underlined} text indicates the second-best.}
\label{m_full_durecdial2.0}
\end{table*}

\begin{table*}[!t]
\centering
\setlength{\tabcolsep}{0.8mm}{
\begin{tabular}{llcccccc}
\toprule
\textbf{Split}                & $m=3$    & \textbf{PPL}$_\downarrow$   & \textbf{Word F1} & \textbf{BLEU-1/2}       & \textbf{DIST-1/2}       & \textbf{Knowledge F1} & \textbf{Failure$_\downarrow$}  \\ \midrule
\multirow{5}{*}{\textbf{ID Test Set}}  & $\delta=0.0$ & 3.98 &  44.61   & 0.413/0.307 & \textbf{0.022}/\textbf{0.088} & 60.70   & 20.72 \\
                     & $\delta=0.1$ &   \textbf{3.92}  & \underline{44.86}   & \underline{0.415}/\underline{0.309} & \textbf{0.022}/\textbf{0.088} & 61.07   & \underline{20.34} \\
                     & $\delta=0.2$ &  \underline{3.93} & \textbf{44.87}   & \textbf{0.416}/\textbf{0.310} & \textbf{0.022}/\underline{0.087} & 61.17   & \textbf{19.77} \\
                     & $\delta=0.3$ &   \underline{3.93}  & 44.84   & \textbf{0.416}/\textbf{0.310} & \textbf{0.022}/\underline{0.087} & \textbf{61.33}   & \textbf{19.77} \\
                     & $\delta=0.4$ &  \underline{3.93}   & 44.84   & \textbf{0.416}/\textbf{0.310} & \textbf{0.022}/\underline{0.087} & \underline{61.29}   & \textbf{19.77} \\ \midrule
\multirow{5}{*}{\textbf{OOD Test Set}} & $\delta=0.0$ &   7.19 & \underline{39.86}   & \textbf{0.365}/\textbf{0.253} & \underline{0.021}/\textbf{0.082} & 49.10   & 25.86 \\
                     & $\delta=0.1$ &  \textbf{7.10}  & \textbf{39.99}   & \textbf{0.365}/\textbf{0.253} & \textbf{0.022}/\underline{0.081} & \underline{49.21}   & \underline{24.30} \\
                     & $\delta=0.2$ &   7.13  &  39.84   & \textbf{0.365}/\textbf{0.253} & \underline{0.021}/0.079 & \textbf{49.27}        & \textbf{23.68} \\
                     & $\delta=0.3$ &   7.13   & 39.82   & \textbf{0.365}/\textbf{0.253} & \underline{0.021}/0.079 & 49.18   & \textbf{23.68} \\
                     & $\delta=0.4$ &   \underline{7.12}  &  39.80   & \textbf{0.365}/\textbf{0.253} & \underline{0.021}/0.079 & \underline{49.21}   & \textbf{23.68} \\ \bottomrule
\end{tabular}}
\caption{Full experiments on the DuRecDial2.0 dataset with different $\delta$ values (when $m=3$). \textbf{Bold} text highlights the best results, while \underline{underlined} text indicates the second-best.}
\label{d_full_durecdial2.0}
\end{table*}

\section{Ablation Study}
We report the ablation experiments on the DuRecDial2.0 dataset here, as illustrated in Table~\ref{ablation_durecdial2.0}.

\begin{table*}[!t]
	\small
	\centering
	\setlength{\tabcolsep}{0.9mm}{
		\begin{tabular}{llllllll}
			\toprule
			\textbf{Split}                & \textbf{Model}  & \textbf{PPL}$_\downarrow$      & \textbf{W. F1} & \textbf{BLEU-1/2}    & \textbf{DIST-1/2}    & \textbf{K. F1} & \textbf{Failure$_\downarrow$} \\ \midrule
			\multirow{5}{*}{\textbf{ID Test Set}}  & \textbf{Ours} & 3.96        & 44.79     & 0.416/0.309 & 0.022/0.088 & 61.11          & 19.77   \\
			& \textbf{w/o IKB} & 3.82$_{\downarrow0.14}$         & 43.33$_{\downarrow1.46}$     & 0.413/0.308$_{\downarrow0.003/0.001}$ & 0.017/0.063$_{\downarrow0.005/0.025}$ & 60.03$_{\downarrow1.08}$          & 25.29$_{\uparrow5.52}$   \\
			& \textbf{w/o CSM}   & 4.01$_{\uparrow0.05}$    &   43.75$_{\downarrow1.04}$     & 0.396/0.288$_{\downarrow0.020/0.021}$ & 0.016/0.059$_{\downarrow0.006/0.029}$ & 53.40$_{\downarrow7.71}$          & 22.43$_{\uparrow2.66}$   \\
			& \textbf{-$\mathcal{F}_k(\mathbf{H}^k)$}   & 4.01$_{\uparrow0.05}$    & 43.89$_{\downarrow0.90}$     & 0.400/0.292$_{\downarrow0.016/0.017}$ & 0.017/0.062$_{\downarrow0.005/0.026}$ & 55.17$_{\downarrow5.94}$          & 19.77$_{\uparrow0.00}$   \\
			& \textbf{-$\mathcal{F}_u(\mathbf{H}^u)$}  & 3.94$_{\downarrow0.02}$    & 44.40$_{\downarrow0.39}$     & 0.408/0.300$_{\downarrow0.008/0.009}$ & 0.017/0.063$_{\downarrow0.005/0.025}$ & 60.24$_{\downarrow0.87}$          & 22.81$_{\uparrow3.04}$   \\
			\midrule
			\multirow{5}{*}{\textbf{OOD Test Set}} & \textbf{Ours}  & 7.27          & 39.61     & 0.363/0.252 & 0.022/0.082 & 49.44          & 23.68   \\
			& \textbf{w/o IKB}    & 6.05$_{\downarrow1.22}$      & 38.05$_{\downarrow1.56}$     & 0.365/0.255$_{\uparrow0.002/0.003}$ & 0.015/0.051$_{\downarrow0.007/0.031}$ & 48.10$_{\downarrow1.34}$          & 26.48$_{\uparrow2.80}$   \\
			& \textbf{w/o CSM}   & 6.10$_{\downarrow1.17}$       & 39.64$_{\uparrow0.03}$     & 0.356/0.244$_{\downarrow0.007/0.008}$ & 0.014/0.048$_{\downarrow0.008/0.034}$ & 42.47$_{\downarrow6.97}$          & 19.94$_{\downarrow3.74}$   \\
			& \textbf{-$\mathcal{F}_k(\mathbf{H}^k)$}    & 6.21$_{\downarrow1.06}$     & 39.41$_{\downarrow0.20}$     & 0.355/0.243$_{\downarrow0.008/0.009}$ & 0.016/0.053$_{\downarrow0.006/0.029}$ & 42.84$_{\downarrow6.60}$          & 19.63$_{\downarrow4.05}$   \\
			& \textbf{-$\mathcal{F}_u(\mathbf{H}^u)$}    & 6.13$_{\downarrow1.14}$   & 39.59$_{\downarrow0.02}$     & 0.363/0.252$_{\downarrow0.000/0.000}$ & 0.015/0.052$_{\downarrow0.007/0.030}$ & 48.67$_{\downarrow0.77}$          & 25.55$_{\uparrow1.87}$   \\
			\bottomrule
	\end{tabular}}
	\caption{Ablation study on the DuRecDial2.0 dataset.
	}
	\label{ablation_durecdial2.0}
\end{table*}

\begin{table}[!t]
	\centering
	\setlength{\tabcolsep}{3mm}{
	\begin{tabular}{lcccc}
		\toprule
		\multicolumn{5}{c}{\textbf{ID   Test Set}}                                      \\ \cmidrule(l){2-5} 
		\textbf{Model}    & \textbf{Pro.} & \textbf{Coh.} & \textbf{App.} & \textbf{Inf.} \\ \midrule
		\textbf{Ours}     & \textbf{3.17}        & \textbf{4.64}      & \underline{4.48}            & \textbf{3.01}            \\
		\textbf{T5-Zh}    & \underline{2.65}        & 4.14      & 4.01            & 2.30            \\
		\textbf{TPDial}   & 2.57        & \underline{4.33}      & \textbf{4.55}            & \underline{2.76}            \\
		\textbf{TRIPDial} & 2.43        & 4.00      & 4.19            & 2.37            \\ \midrule
		\multicolumn{5}{c}{\textbf{OOD   Test Set}}                                     \\ \midrule
		\textbf{Ours}     & \textbf{3.01}        & \textbf{4.73}      & \underline{4.47}            & \textbf{2.87}            \\
		\textbf{T5-Zh}    & \underline{2.49}        & 4.13      & 3.98            & 2.09            \\
		\textbf{TPDial}   & 2.18        & 4.09      & 3.74            & 2.16            \\
		\textbf{TRIPDial} & 2.40        & \underline{4.22}      & \textbf{4.48}            & \underline{2.48}            \\ \bottomrule
	\end{tabular}}
	\caption{LLM-as-a-judge scores of advanced models on the DuRecDial dataset. "Pro." = Proactivity, "Coh." = Coherence, "App." = Appropriateness, "Inf." = Informativeness. \textbf{Bold} text indicates the best results, and \underline{underlined} text indicates the second-best.}
	\label{llm_as_judge_plm_durecdial}
\end{table}

\begin{table}[!t]
	\centering
	\setlength{\tabcolsep}{3mm}{
	\begin{tabular}{lcccc}
		\toprule
		\multicolumn{5}{c}{\textbf{ID   Test Set}}                                      \\ \cmidrule(l){2-5} 
		\textbf{Model}    & \textbf{Pro.} & \textbf{Coh.} & \textbf{App.} & \textbf{Inf.} \\ \midrule
		\textbf{Ours}     & \textbf{2.90}        & \textbf{4.64}      & \textbf{4.53}            & \textbf{2.75}            \\
		\textbf{T5-Flan}  & \underline{2.50}        & \underline{4.35}      & \underline{4.08}            & 2.20            \\
		\textbf{TPDial}   & 2.31        & 4.15      & 4.00            & 2.13            \\
		\textbf{TRIPDial} & 2.33        & 4.21      & 3.98            & \underline{2.25}            \\ \midrule
		\multicolumn{5}{c}{\textbf{OOD   Test Set}}                                     \\ \midrule
		\textbf{Ours}     & \textbf{3.03}        & \textbf{4.53}      & \textbf{4.28}            & \textbf{2.58}            \\
		\textbf{T5-Flan}  & \underline{2.60}        & \underline{4.25}      & \underline{3.77}            & 2.10            \\
		\textbf{TPDial}   & 2.20        & 3.95      & 3.43            & 1.73            \\
		\textbf{TRIPDial} & 2.24        & 3.95      & 3.63            & \underline{2.15}            \\ \bottomrule
	\end{tabular}}
	\caption{LLM-as-a-judge scores of advanced models on the DuRecDial2.0 dataset. "Pro." = Proactivity, "Coh." = Coherence, "App." = Appropriateness, "Inf." = Informativeness. \textbf{Bold} text indicates the best results, and \underline{underlined} text indicates the second-best.}
	\label{llm_as_judge_plm}
\end{table}

\begin{table}[!t]
	\centering
	\setlength{\tabcolsep}{2.5mm}{
	\begin{tabular}{lcccc}
		\toprule
		\multicolumn{5}{c}{\textbf{ID   Test Set}}                                     \\ \cmidrule(l){2-5} 
		\textbf{Model}   & \textbf{Pro.} & \textbf{Coh.} & \textbf{App.} & \textbf{Inf.} \\ \midrule
		\textbf{Ours}$_{\text{0.3B}}$    & \underline{2.85}        & \underline{4.41}      & 4.29            & \underline{2.55}            \\
		\textbf{LLaMA}$_{\text{1B}}^\diamondsuit$ & 2.23        & 4.26      & 4.16            & 2.29            \\
		\textbf{LLaMA}$_{\text{3B}}^\diamondsuit$ & 2.32        & 4.29      & 4.31            & 2.33            \\
		\textbf{Qwen}$_{\text{3B}}^\diamondsuit$  & 2.44        & 4.35      & \underline{4.39}            & 2.40            \\
		\textbf{LLaMA}$_{\text{8B}}^\heartsuit$ & 3.36        & 4.40      & 4.35            & 3.44            \\
		\textbf{Qwen}$_{\text{14B}}^\heartsuit$ & \textbf{4.21}        & \textbf{4.81}      & 4.71            & 3.76            \\
		\textbf{Qwen}$_{\text{32B}}^\heartsuit$ & 4.06        & \textbf{4.81}      & \textbf{4.76}            & \textbf{4.05}            \\ \midrule
		\multicolumn{5}{c}{\textbf{OOD   Test Set}}                                    \\ \midrule
		\textbf{Ours}$_{\text{0.3B}}$    & \underline{2.66}        & \underline{4.48}      & 4.33            & \underline{2.40}            \\
		\textbf{LLaMA}$_{\text{1B}}^\diamondsuit$ & 2.06        & 4.30      & 4.36            & 2.26            \\
		\textbf{LLaMA}$_{\text{3B}}^\diamondsuit$ & 2.19        & 4.40      & 4.53            & 2.37            \\
		\textbf{Qwen}$_{\text{3B}}^\diamondsuit$  & 2.34        & 4.40      & \underline{4.62}            & 2.37            \\
		\textbf{LLaMA}$_{\text{8B}}^\heartsuit$ & 3.09        & 4.45      & 4.42            & 3.35            \\
		\textbf{Qwen}$_{\text{14B}}^\heartsuit$ & \textbf{4.15}        & 4.80      & 4.78            & 3.77            \\
		\textbf{Qwen}$_{\text{32B}}^\heartsuit$ & 3.85        & \textbf{4.81}      & \textbf{4.81}            & \textbf{3.98}            \\ \bottomrule
	\end{tabular}}
	\caption{LLM-as-a-judge scores of LLMs on DuRecDial. "Pro." = Proactivity, "Coh." = Coherence, "App." = Appropriateness, "Inf." = Informativeness. Models labeled $^\diamondsuit$ denote fine-tuning, while models labeled $^\heartsuit$ denote the prompt-based method. \textbf{Bold} text indicates the best-performing model; \underline{underlined} text indicates the best-performing model among the fine-tuning methods.
	}
	\label{llm_as_judge_llm_durecdial_zh}
\end{table}

\begin{table}[!t]
	\centering
	\setlength{\tabcolsep}{2.5mm}{
	\begin{tabular}{lcccc}
		\toprule
		\multicolumn{5}{c}{\textbf{ID   Test Set}}                                     \\ \cmidrule(l){2-5} 
		\textbf{Model}   & \textbf{Pro.} & \textbf{Coh.} & \textbf{App.} & \textbf{Inf.} \\ \midrule
		\textbf{Ours}$_{\text{0.3B}}$    & \underline{2.45}        & \underline{4.41}      & 4.30            & \underline{2.16}            \\
		\textbf{LLaMA}$_{\text{1B}}^\diamondsuit$ & 2.12        & 4.32      & 4.25            & 2.07            \\
		\textbf{LLaMA}$_{\text{3B}}^\diamondsuit$ & 2.08        & 4.23      & 4.18            & 1.98            \\
		\textbf{Qwen}$_{\text{3B}}^\diamondsuit$  & 2.21        & 4.30      & \underline{4.31}            & 2.05            \\
		\textbf{LLaMA}$_{\text{8B}}^\heartsuit$ & 3.74        & \textbf{4.82}      & 4.73            & 4.17            \\
		\textbf{Qwen}$_{\text{14B}}^\heartsuit$ & \textbf{3.86}        & \textbf{4.91}      & \textbf{4.85}            & 3.83            \\
		\textbf{Qwen}$_{\text{32B}}^\heartsuit$ & \textbf{3.86}        & \textbf{4.91}      & \textbf{4.85}            & 4.01            \\ \midrule
		\multicolumn{5}{c}{\textbf{OOD   Test Set}}                                    \\ \midrule
		\textbf{Ours}$_{\text{0.3B}}$    & \underline{2.71}        & \underline{4.33}      & 3.98            & \underline{1.98}            \\
		\textbf{LLaMA}$_{\text{1B}}^\diamondsuit$ & 2.31        & \underline{4.33}      & 3.97            & 1.89            \\
		\textbf{LLaMA}$_{\text{3B}}^\diamondsuit$ & 2.32        & 4.29      & 4.07            & 1.89            \\
		\textbf{Qwen}$_{\text{3B}}^\diamondsuit$  & 2.35        & 4.29      & \underline{4.09}            & 1.88            \\
		\textbf{LLaMA}$_{\text{8B}}^\heartsuit$ & 3.78        & 4.82      & 4.69            & 4.03            \\
		\textbf{Qwen}$_{\text{14B}}^\heartsuit$ & 4.02        & \textbf{4.92}      & 4.83            & 3.83            \\
		\textbf{Qwen}$_{\text{32B}}^\heartsuit$ & \textbf{4.07}        & 4.91      & \textbf{4.86}            & \textbf{4.13}            \\ \bottomrule
	\end{tabular}}
	\caption{LLM-as-a-judge scores of LLMs on DuRecDial2.0. "Pro." = Proactivity, "Coh." = Coherence, "App." = Appropriateness, "Inf." = Informativeness. Models labeled $^\diamondsuit$ denote fine-tuning, while models labeled $^\heartsuit$ denote the prompt-based method. \textbf{Bold} text indicates the best-performing model; \underline{underlined} text indicates the best-performing model among the fine-tuning methods.
	}
	\label{llm_as_judge_llm}
\end{table}

\section{Human and LLM-as-a-Judge Evaluations}
\label{human_evaluate_appendix}
Since automatic metrics cannot fully capture the quality of system utterances, they are complemented by additional evaluations: (i) we conduct pairwise human evaluations comparing our framework with the strong backbone models T5-Zh and T5-Flan on the DuRecDial and DuRecDial2.0 datasets, respectively; (ii) an LLM-as-a-judge approach is employed to further assess the performance of our framework against the baseline models.

\begin{figure}[!t]
	\small
    \centering
   \includegraphics[width=\columnwidth]{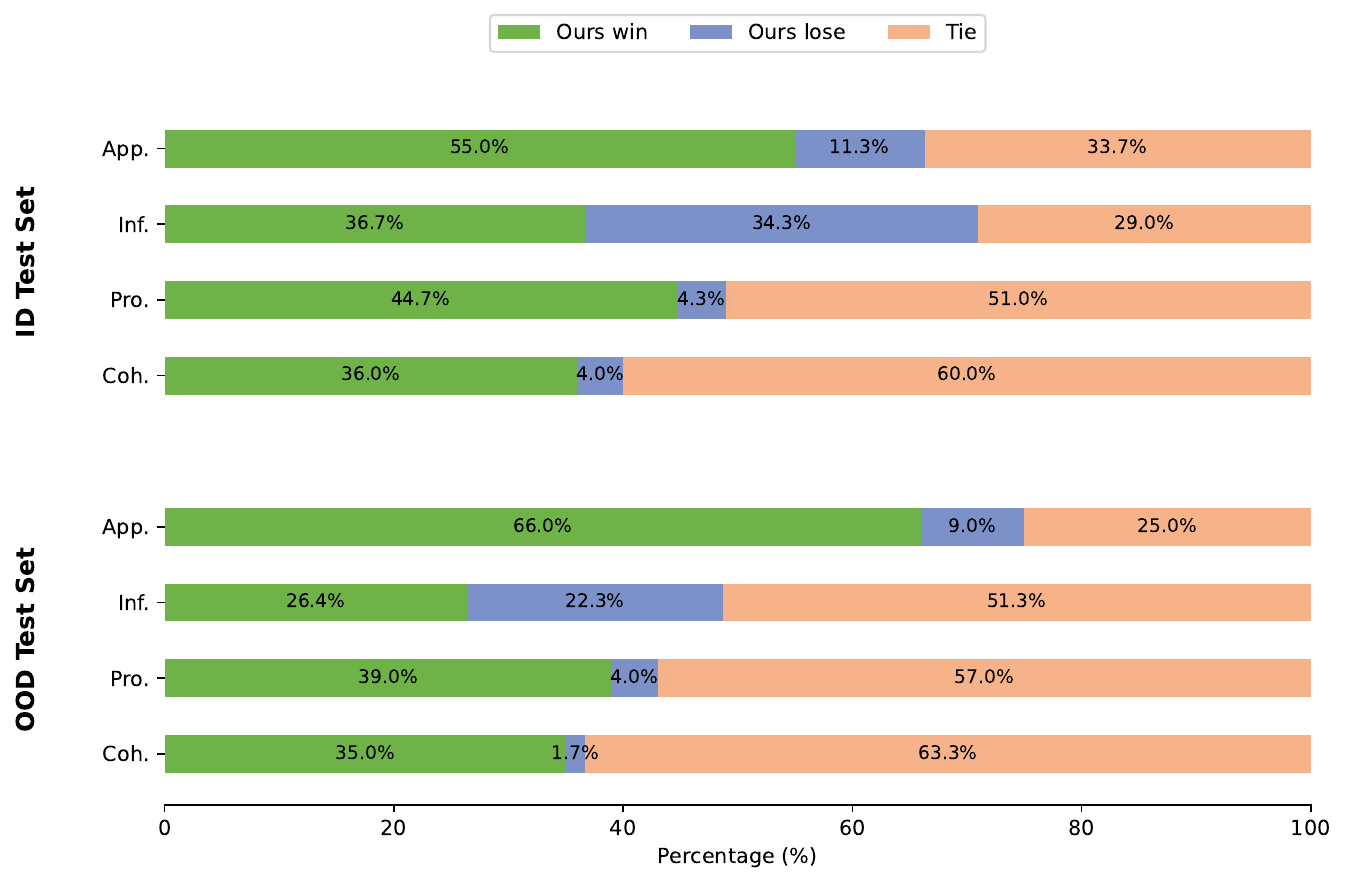}
   \caption{
  Human evaluation on the DuRecDial dataset (our framework vs. T5-Zh). "App." denotes "appropriateness," "Inf." denotes "informativeness," "Pro." denotes "proactivity," and "Coh." denotes "coherence."
}
  \label{human_evaluate_durecdial}
\end{figure}

For pairwise human evaluation, we follow both turn-level and dialogue-level evaluations. For the turn-level evaluation, we consider two aspects: \textbf{appropriateness (App.)} and \textbf{informativeness (Inf.)}. (1) Appropriateness measures whether the system utterance aligns with the current intent keyword; (2) informativeness evaluates whether the system utterance incorporates relevant user profiles and domain knowledge. For the dialogue-level evaluation, we consider two aspects: \textbf{proactivity (Pro.)} and \textbf{coherence (Coh.)}. (1) Proactivity evaluates whether the system can proactively introduce new keyword–topics; (2) coherence evaluates the overall fluency and naturalness of the generated dialogue. We randomly sample 500 utterances and evaluate each utterance independently, without considering the overall dialogue context. For dialogue-level evaluation, although 500 utterances are also sampled, the entire dialogue process is considered, i.e., whether the dialogue progresses coherently. All samples are labeled by three graduate students with backgrounds in natural language processing. The labels "Win," "Tie," and "Lose" indicate whether our framework performs better, comparably, or worse, respectively, compared with the strong backbone models T5-Zh and T5-Flan. We average the labels provided by the three graduate students as the final results and report the corresponding percentages in Figures~\ref{human_evaluate_durecdial} and~\ref{human_evaluate_durecdial2.0}\footnote{We collect responses from the forms using Tencent Questionnaire and perform unified calculations to derive the final results.}. Figure~\ref{human_evaluate_durecdial} further validates our framework's substantial improvements in proactivity, fluency, and informativeness. In the evaluations of conversational coherence and proactivity, our framework demonstrates a clear advantage: on the ID test set, it significantly outperforms the baseline's 4.0\% and 4.3\% with win rates of 36.0\% and 44.7\%, respectively; on the OOD test set, it also maintains a strong lead with win rates of 35.0\% and 39.0\%. More importantly, when facing the OOD test set, our framework achieves an impressive win rate of 66.0\% in the appropriateness metric (compared to the baseline's 9.0\%), while consistently maintaining a net advantage in informativeness. This demonstrates that the synergy between CSM and IKB can advance the dialogue trajectory and enrich contextual content in a highly natural manner, effectively avoiding the mechanical and rigid utterances produced by traditional systems in pursuit of their targets.

\begin{figure*}[!t]
	\centering
	\includegraphics[width=0.91\textwidth]{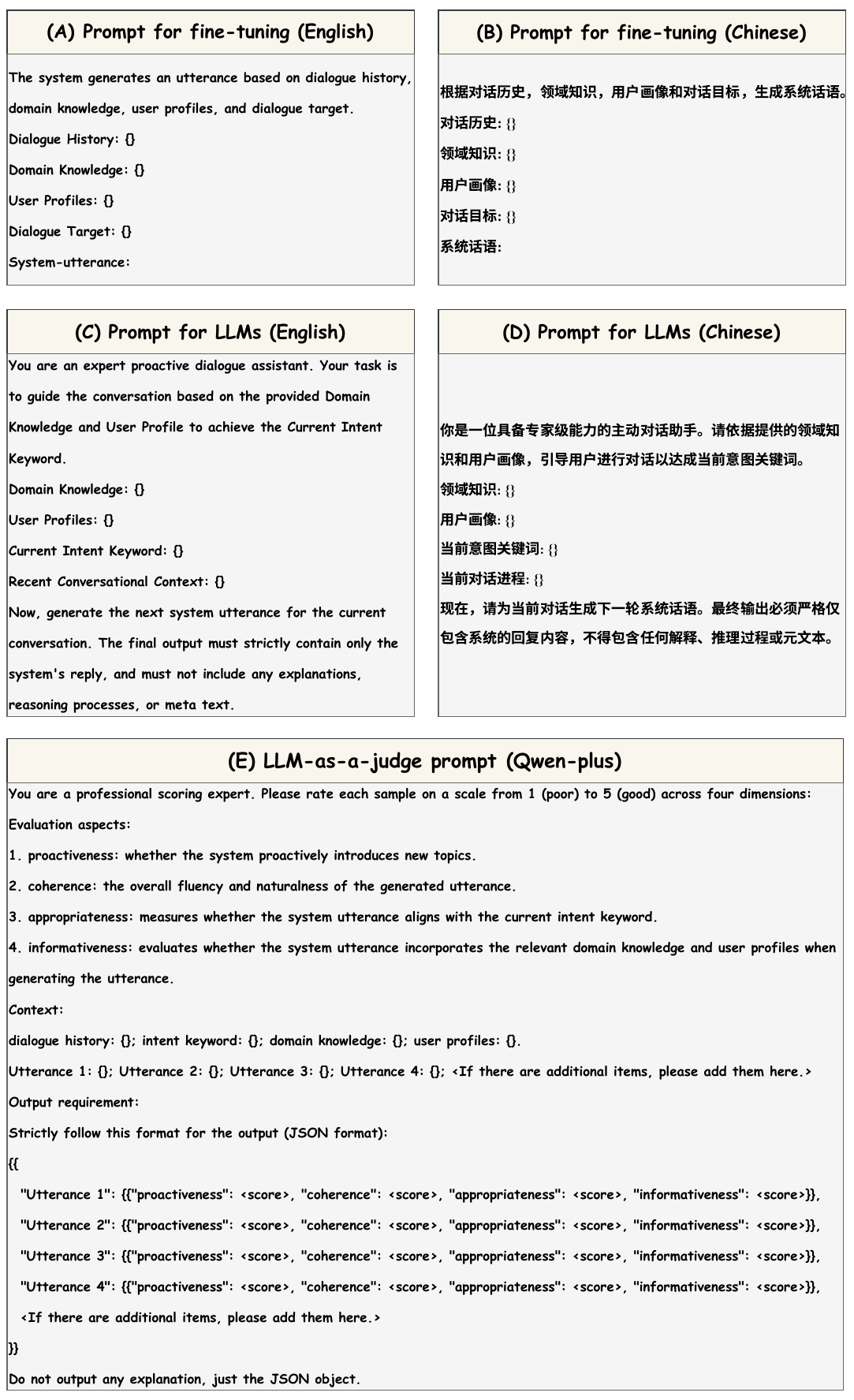}
	\caption{
		Our using prompt.
	}
	\label{appendix_prompt_fig}
\end{figure*}

For the LLM-as-a-judge approach, the prompt template used in this work is shown in Figure~\ref{appendix_prompt_fig}(E)\footnote{We use the latest Qwen-plus LLM.}, and the corresponding results are presented in Tables~\ref{llm_as_judge_plm_durecdial}, \ref{llm_as_judge_plm}, \ref{llm_as_judge_llm_durecdial_zh}, and \ref{llm_as_judge_llm}. 

Tables~\ref{llm_as_judge_plm_durecdial} and~\ref{llm_as_judge_plm} present the automatic evaluation results based on the LLM-as-a-judge, compared with advanced models on the DuRecDial and DuRecDial2.0 datasets, respectively. These results are highly consistent with the objective metrics in Table~\ref{mainresult_plm} from the main experiment, further solidifying the core advantages of our framework in improving proactivity, coherence, and informativeness. In direct comparisons with advanced models (such as TPDial and TRIPDial), our framework achieves state-of-the-art performance across almost all evaluation dimensions on the ID and OOD test sets of both the DuRecDial and DuRecDial2.0 datasets. In particular, for proactivity and coherence—two metrics directly measuring the naturalness of interaction—our framework significantly and consistently outperforms strong baseline models. This demonstrates that our framework provides multi-step dynamic look-ahead guidance, enabling smooth and natural topic transitions and effectively avoiding the abrupt utterances produced by traditional systems in pursuit of their targets. Meanwhile, the absolute lead in informativeness further indicates that by introducing scenario bias, our framework successfully integrates user profiles and domain knowledge into the utterance generation process, greatly enriching the contextual information of the dialogue.

Tables~\ref{llm_as_judge_llm_durecdial_zh} and~\ref{llm_as_judge_llm} present an in-depth comparison of the performance of our lightweight framework (Ours 0.3B) with LLMs of various sizes on the DuRecDial and DuRecDial2.0 datasets, respectively. Their conclusions align with the findings in Table~\ref{mainresult_llm} from the main experiment, fully demonstrating the superior architectural performance of the proposed mechanism even when compared with the large number of parameters in the LLM family. Among all LLMs employing efficient parameter fine-tuning (including LLaMA$_\text{1B/3B}$ and Qwen$_\text{3B}$), our framework, with only 0.3B parameters, consistently ranks highest in proactivity, coherence, appropriateness, and informativeness on the ID and OOD test sets of both datasets. While large-parameter prompt-based models (such as Qwen$_\text{14B/32B}$) achieve higher absolute scores due to their extensive internal knowledge (since prompt-based methods typically produce longer utterances, the LLM-as-a-judge usually assigns higher scores), as shown in our main experiment (Table~\ref{mainresult_llm}), these models often compromise dialogue quality (W. F1) and produce abrupt topic transitions. In contrast, our framework achieves an optimal balance between generation quality, proactive guidance, and natural interaction, while maintaining minimal computational overhead and a lightweight footprint. This not only demonstrates the substantial potential of our framework in resource-constrained scenarios but also highlights that integrating conversational scenario modeling with higher-level, flexible intent keyword guidance offers an effective path toward bringing target-guided proactive dialogue systems closer to real-world human interaction.

\begin{figure*}[!t]
\centering
  \includegraphics[width=\textwidth]{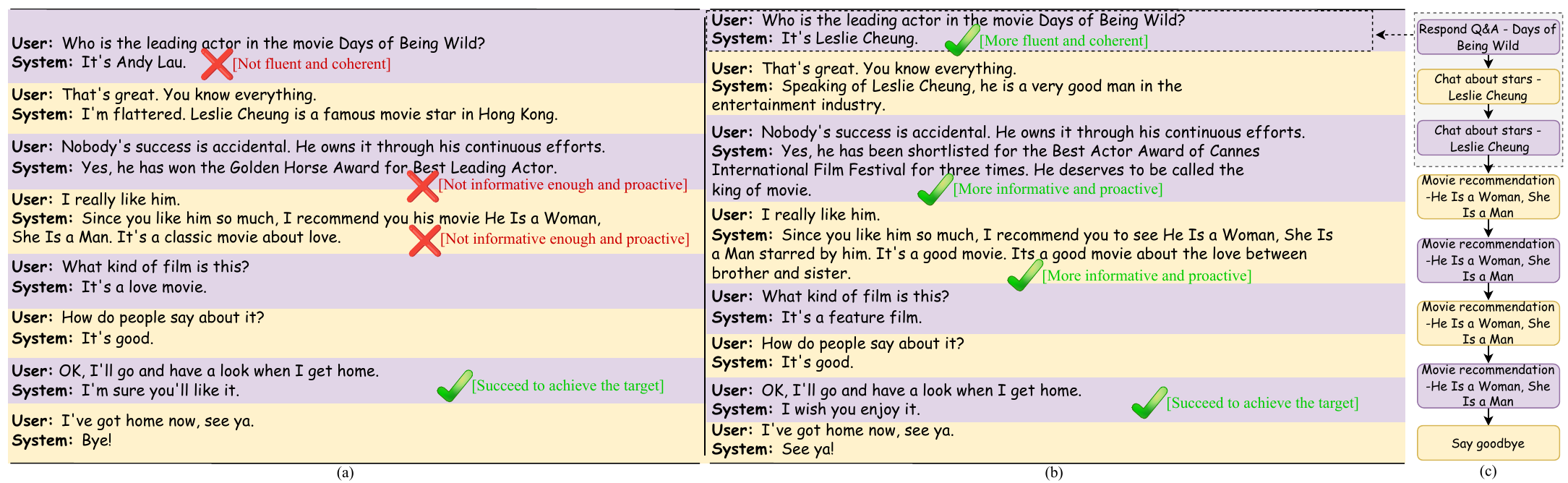}
  \caption{
  Case study (OOD test set). (a) Results of T5-Flan; (b) Results of our framework; (c) Dialogue-level intent keyword transitions across turns.
}
  \label{case2}
\end{figure*}

\begin{table}[!t]
\small
\centering
\begin{tabular}{|p{0.9\linewidth}|}
\hline
\begin{minipage}[t]{\linewidth}
Name: \textit{Xinling Chen} \\
Age Range: \textit{26-35} \\
Gender: \textit{Female} \\
Residence: \textit{Haikou} \\
Occupation: \textit{Employed} \\
\textbf{Accepted Celebrities: \textit{Hsu Chi; Leehom Wang}} \\
\textbf{Accepted Music: \textit{A Simple Song; Heroes of Earth; All the Things You Never Knew; Secret Garden; Bosom Friend}} \\
\textbf{Rejected Music: \textit{KISS GOODBYE; That Year; Change Me}} \\
Accepted POI: \textit{One Plus One Northeastern Chinese Dishes and Dumplings (Renmin Avenue Store)} \\
etc.
\end{minipage}
 \\ \hline
\end{tabular}
\caption{Visualization of user profile modeling $\mathcal{F}_u(\mathbf{H}^u)$ in the Case Study (ID). \textbf{Bold} highlights the information identified as relevant by the user profile modeling.}
\label{case1_up}
\end{table}

\begin{table}[!t]
\small
\centering
\begin{tabular}{|p{0.9\linewidth}|}
\hline
\begin{minipage}[t]{\linewidth}
\textit{<Leehom Wang, Birthday, 1976 - 5 - 17>} \\
\textit{<Leehom Wang, Achievement, The Most Popular Male Singer of MTV Asian Music in Taiwan>} \\
\textit{<Leehom Wang, Achievement, Best Male Singer of Global Chinese Golden Chart>}\\
\textit{<Leehom Wang, Achievement, the 15th Global Chinese Music Award for Best Male Singer in Hong Kong and Taiwan>} \\
\textit{<Leehom Wang, Awards, The Chinese Film Media Award for Most Popular Actor voted by the audience>} \\
\textit{<Leehom Wang, Sings, The Sun Washed by Spring Rain>} \\
etc.
\end{minipage}
 \\ \hline
\end{tabular}
\caption{
Visualization of domain knowledge modeling $\mathcal{F}_k(\mathbf{H}^k)$ in the Case Study (ID).
}
\label{case1_kg}
\end{table}

\begin{table}[!t]
\small
\centering
\begin{tabular}{|p{0.9\linewidth}|}
\hline
\begin{minipage}[t]{\linewidth}
Name: \textit{Pingshan Han} \\
Age Range: \textit{18-25} \\
Gender: \textit{Female} \\
Residence: \textit{Nantong} \\
Occupation: \textit{Employed} \\
Accepted Celebrities: \textit{Leslie Cheung} \\
\textbf{Accepted Movies: \textit{Days of Being Wild; Moonlight Express; Happy Together}} \\
Rejected Movies: \textit{Double Tap} \\
Accepted Food: \textit{Marinated Fish} \\
\textbf{Accepted POI: \textit{Lahuangshang Spicy Pot Roasted Fish}} \\
Favorite News: \textit{Leslie Cheung's News} \\
etc.
\end{minipage}
 \\ \hline
\end{tabular}
\caption{Visualization of user profile modeling $\mathcal{F}_u(\mathbf{H}^u)$ in the Case Study (OOD). \textbf{Bold} highlights the information identified as relevant by the user profile modeling.}
\label{case2_up}
\end{table}

\begin{table}[!t]
\small
\centering
\begin{tabular}{|p{0.9\linewidth}|}
\hline
\begin{minipage}[t]{\linewidth}
\textit{<Leslie Cheung, Stars, Days of Being Wild>} \\
\textit{<Days of Being Wild, Director, Wong Kar Wai>} \\
\textit{<Days of Being Wild, Reputation, The reputation is good. Its rating is in the top 5 this year. Its rating is in the top 10 this year>} \\
\textit{<Leslie Cheung, Achievement, Three times shortlisted for the Best Actor Award in the Cannes International Film Festival>} \\
\textit{<Leslie Cheung, Achievement, introduced in Encyclopedia Britannica>} \\
\textit{<Leslie Cheung, Stars, He Is a Woman, She Is a Man>} \\
etc.
\end{minipage}
 \\ \hline
\end{tabular}
\caption{
Visualization of domain knowledge modeling $\mathcal{F}_k(\mathbf{H}^k)$ in the Case Study (OOD).
}
\label{case2_kg}
\end{table}

\begin{table*}[!t]
	\centering
	\setlength{\tabcolsep}{0.9mm}{
	\begin{tabular}{lcccc}
		\toprule
		\multicolumn{5}{c}{\textbf{DuRecDial   Dataset}}                                                                                               \\ \midrule
		\multirow{2}{*}{\textbf{Model}} & \multicolumn{2}{c|}{\textbf{ID   Test Set}}                              & \multicolumn{2}{c}{\textbf{OOD   Test Set}}         \\ \cmidrule(l){2-5} 
		& \textbf{F1   (Keyword-type)} & \multicolumn{1}{c|}{\textbf{F1   (Keyword-topic)}} & \textbf{F1   (Keyword-type)} & \textbf{F1   (Keyword-topic)} \\ \midrule
		\textbf{LLaMA}                  & 98.23               & \multicolumn{1}{c|}{97.43}                & 98.45               & 97.31                \\ \midrule
		\multicolumn{5}{c}{\textbf{DuRecDial2.0   Dataset}}                                                                                            \\ \midrule
		\textbf{LLaMA}                  & 98.42               & \multicolumn{1}{c|}{96.68}                & 98.37               & 92.47                \\ \bottomrule
	\end{tabular}}
	\caption{Intent keyword prediction performance.}
	\label{intent_keyword_prediction_preformance}
\end{table*}

\section{Case Study}
\label{case_study}
To more clearly and intuitively evaluate the performance of our framework, we conduct another case study on the ID and OOD test sets, as shown in Figure~\ref{case2}. Notably, the case study adopts the hard mode, enabling a clearer comparison with T5-Flan. Using examples from the OOD test set, Figure~\ref{case2} further validates that even when faced with target topics not encountered during training, our framework significantly improves dialogue proactivity, fluency, and informativeness. At the beginning of the dialogue, the baseline model T5-Flan makes a factual error (\textit{It's Andy Lau.}) and subsequently offers only general terms (\textit{It's a classic movie about love.}), revealing its shortcomings in knowledge coverage and planning in unseen domains. Conversely, thanks to the strong generalization ability of our framework (as revealed by the ablation experiments, where removing CSM leads to a sharp drop in knowledge utilization K. F1 on the OOD test set), our framework not only accurately generates (\textit{It's Leslie Cheung.}) but also proactively introduces deeper domain knowledge such as (\textit{shortlisted for the Best Actor Award of Cannes}), greatly enriching the dialogue. Meanwhile, IKB continues to demonstrate outstanding dynamic guidance on the OOD test set, ensuring natural and coherent logical transitions. This aligns with its decisive role in reducing the high failure rate of OOD targets observed in the ablation experiments, further confirming that our framework effectively bridges the gap between target-guided dialogue systems and natural real-world interactions.

At the same time, we examine the content emphasized by conversational scenario modeling—namely, user profile modeling $\mathcal{F}_u(\mathbf{H}^u)$ and domain knowledge modeling $\mathcal{F}_k(\mathbf{H}^k)$—which are jointly employed to capture conversational scenario bias in this work (see \S\ref{CSM_section}). We examine the user profile and domain knowledge items associated with tokens that receive higher modeling probabilities within a conversation. For the example in Figure~\ref{case1}, the information highlighted in the user profiles and domain knowledge is presented in Table~\ref{case1_up} and Table~\ref{case1_kg}, respectively. Similarly, for the example in Figure~\ref{case2}, the information highlighted in the user profiles and domain knowledge is presented in Table~\ref{case2_up} and Table~\ref{case2_kg}, respectively. For user profiles, \textbf{bold} is used to mark the emphasized information. For domain knowledge, since the domain knowledge is pre-processed and all items are considered by the domain knowledge modeling, we only visualize the items relevant to the current conversation. 

As shown in Table~\ref{case1_up}, in this conversation, the user profile modeling focuses on \textit{Accepted Celebrities} as well as items related to \textit{Accepted Music} and \textit{Rejected Music}. These indicate the celebrity and music categories that the user is interested in; however, this does not necessarily imply that the user must accept or reject a specific item, as the actual dialogue flow must also be considered. Overall, it can be generally observed that users discuss celebrities first, followed by music-related topics. From Table~\ref{case1_kg}, we can observe that all triples beginning with \textit{Leehom Wang} are highlighted. His birthday (\textit{1976–5–7}) contributes to the first turn; his achievements and awards play a role in the third and fourth turns; and his song \textit{The Sun Washed by Spring Rain} informs the fourth turn. Together, these elements provide additional information that enriches the dialogue, thereby enhancing the fluency and informativeness of the utterances. As shown in Table~\ref{case2_up}, we observe that, in this user profile, the modeling highlights \textit{Accepted Movies} and \textit{Accepted POI}, even though no point-of-interest information appears in the conversation. In contrast, the \textit{Accepted Celebrities: Leslie Cheung}—which is directly relevant to the dialogue—is not emphasized in the user profile. These observations suggest that, on the one hand, the content identified by the modeling is meaningful only when interpreted within the specific dialogue context; on the other hand, on the OOD test set, keyword-topics absent from the training data negatively affect user profile modeling. From Table~\ref{case2_kg}, we observe that all triples beginning with \textit{Leslie Cheung} are highlighted. His starred movie \textit{Days of Being Wild} plays a role in the first turn, and \textit{He Is a Woman, She Is a Man} contributes to the fourth turn; his achievements and awards inform the second turn. Moreover, the \textit{Direct} and \textit{Reputation} aspects of \textit{Days of Being Wild} further enrich the conversation, as evidenced in the second turn. 

Whether in the user profiles or domain knowledge, much of the information is either overlooked or, although emphasized, may not be fully reflected in a single conversation. This is expected, as such information assumes a more meaningful role when combined with intent keywords and the specific dialogue context. Items previously ignored may be highlighted due to the intent keywords, whereas items that were previously focused but are irrelevant to the current conversation will be disregarded to avoid adverse effects, thereby generating more proactive, fluent, and informative utterances. It is worth noting that both user profile modeling and domain knowledge modeling operate at the conversational level and cannot be applied to individual dialogue turns. The above descriptions are intended to illustrate their respective roles. At the conversational level, user profile and domain knowledge modeling (i.e., conversational scenario modeling) jointly influence and enrich the generated utterances, enabling precise initiative and enhanced user engagement.

To sum up, user profile modeling renders the utterance more consistent with user profiles, thereby increasing lexical diversity, whereas domain knowledge modeling aligns the utterance with domain knowledge, improving knowledge utilization and indirectly enhancing user engagement. Taken together with the preceding ablation studies (\S\ref{ablation_study_main}), these examples further validate the effectiveness of conversational scenario modeling for utterance generation, demonstrating substantial gains across multiple aspects—particularly in lexical diversity and user engagement—while enabling precise initiative.

\section{Prompting Template}
\label{appendix_prompt}
This section presents the prompt templates used, specifically the templates for fine-tuning LLMs. Figures~\ref{appendix_prompt_fig}(A) and \ref{appendix_prompt_fig}(B) correspond to the DuRecDial and DuRecDial2.0 datasets, respectively. In the prompt-based LLM approach, Figures~\ref{appendix_prompt_fig}(C) and \ref{appendix_prompt_fig}(D) are used for the DuRecDial and DuRecDial2.0 datasets. Notably, the prompt-based method also requires an intent keyword field. We employ LLaMA-8B as the predictor, and the corresponding performance is reported in Table~\ref{intent_keyword_prediction_preformance}. We adopt the F1 score (F1), which measures the micro-averaged precision and recall of the predicted keyword types and keyword topics. As shown in Table~\ref{intent_keyword_prediction_preformance}, the prediction performance of intent keywords is strong, indicating that the prompt-based LLM method compared in this paper is reliable (see Table~\ref{mainresult_llm}).

\end{document}